\documentclass[wcp]{jmlr}

\usepackage{longtable}

\usepackage{booktabs}

\title[Short Title]{Byzantine-Robust Federated Learning via Credibility Assessment on Non-IID Data}

\author{\Name{Kun Zhai} \Email{1930800@tongji.edu.cn}
  \AND
  \Name{Qiang Ren} \Email{rqfzpy@tongji.edu.cn}
  \AND
  \Name{Junli Wang} \Email{junliwang@tongji.edu.cn}
  \AND
  \Name{Chungang Yan} \Email{yanchungang@tongji.edu.cn}\\
  \\
  \addr{Key Laboratory of Embedded System and Service Computing (Tongji University), Ministry of Education, Shanghai 201804, China.}
}
\usepackage{indentfirst}
\usepackage[mathscr]{eucal}
\usepackage{algorithm2e}
\usepackage{multicol}
\usepackage{multirow}
\usepackage{caption}
\usepackage{graphicx}
\usepackage{float}
\usepackage{booktabs}
\usepackage{amsmath,bm}
\usepackage{diagbox}
\usepackage{setspace}

\begin{document}

\maketitle

\begin{abstract}
Federated learning is a novel framework that enables resource-constrained edge devices to jointly learn a model, which solves the problem of data protection and data islands. However, standard federated learning is vulnerable to Byzantine attacks, will cause the global model to be manipulated by the attacker or fail to converge. On non-iid data, the current methods are not effective in defensing against Byzantine attacks. In this paper, we propose a Byzantine-robust framework for federated learning via credibility assessment on non-iid data (\emph{BRCA}). First, a credibility assessment method is designed to detect Byzantine attacks by combing adaptive anomaly detection model and data verification. Specially, an adaptive mechanism is incorporated into the anomaly detection model for the training and prediction of the model. Final, a unified update algorithm is given to guarantee that the global model has a consistent direction. On non-iid data, our experiments demonstrate that the \emph{BRCA} is more robust to Byzantine attacks compared with conventional methods.
\end{abstract}

\section{Introduction}
With the popularity of smartphones, wearable devices, intelligent home appliances, and autonomous driving, these distributed devices generate a great number of data every day. In order to make full and effective use, these data are usually sent to the data center for processing and learning tasks. However, the direct collection of personal information from distributed devices has caused people to worry about personal privacy \citep{zhou2018security}. At the same time, as the computing power of these mobile devices increases, it is attractive to store data locally while completing related computing tasks. Federated learning is a distributed machine learning framework that allows multiple parties to collaboratively train a model without sharing raw data~\citep{mcmahan2017communication}; ~\citep{zhao2018federated}, which has attracted significant attention from industry and academia recently. Although federated learning has essential significance and advantages in protecting user privacy, it also faces many challenges. 

First of all, due to the distributed nature of federated learning, it is vulnerable to Byzantine attacks. Notably, it has been shown that, with just one Byzantine client, the whole federated optimization algorithm can be compromised and may lead to failure to converge ~\citep{mhamdi2018hidden}. Especially when the training data is not independent and identically distributed (non-iid), the difficulty of defense against Byzantine attacks is increased and it is difficult to guarantee the convergence of the model ~\citep{so2020byzantine}.

\par Methods for defending against Byzantine attacks in federated learning have been extensively studied, including coordinate-wise trimmed mean \citep{yin2018byzantine}, the coordinate-wise median \citep{chen2019distributed}, \citep{yang2019byzantine}, the geometric median \citep{chen2017distributed}, \citep{pillutla2019robust}, and distance-based methods Krum \citep{blanchard2017machine}, BREA \citep{so2020byzantine}, Bulyan \citep{mhamdi2018hidden}. In addition to the above methods based on statistical knowledge, \cite{li2020learning} proposes a new idea based on anomaly detection to complete the detection of Byzantine clients in the learning process. 

\par The above methods can effectively defend against Byzantine attacks to a certain extent, but there are also some shortcomings. First, the methods based on statistical knowledge have high computational complexity, and also their defense abilities are weakened due to the non-iid characteristics of client data in federated learning. Second, for the anomaly detection algorithm \citep{li2020learning}, there is a premise that the detection model should be trained on the test data set. Obviously, the premise hypothesis cannot be realized in practical applications because it is difficult for us to get such a data set, which can cover almost all data distributions.

\par In this paper, we propose a new method which makes a trade-off between client privacy and model performance. Each client needs to share some data with the server. Unlike FedAvg \citep{mcmahan2017communication}, we use credibility score as the weight of model aggregation, not the sample size. The credibility score of each client is obtained by integrating the verification score and the detection score. The former is calculated by sharing data. The latter is obtained by the adaptive anomaly detection model, which is pre-trained on the source domain, and dynamically fine-tuned on the target domain. Finally, we complete the \emph{unified update} of the global model on the shared data to enable the global model to have a consistent direction.

\par For image classification tasks, we use logistic regression and convolutional neural networks to evaluate our algorithms on the Mnist and Cifar10. In these experiments, our algorithm effectively defends against Byzantine attacks and improves the performance of federated learning. 

The main contributions of this paper are:

\par $\bullet$ We propose a new federated learning framework (\emph{BRCA}) which combines \emph{credibility assessment} and \emph{unified update}. \emph{BRCA} not only effectively defends against Byzantine attacks, but also reduces the impact of non-iid data on the aggregated global model. 

\par $\bullet$ The \emph{credibility assessment} combing anomaly detection and data verification effectively detects Byzantine attacks on non-iid data. 
\par $\bullet$ By incorporating an adaptive mechanism and transfer learning into the anomaly detection model, the anomaly detection model can dynamically improve detection performance. Moreover, its pre-training no longer relies on the test data set. 

\par $\bullet$  We construct four different data distributions for each data set, and conduct experiments on them to study the influence of data distribution on defense methods against Byzantine attacks.

\section{Related Work}

\emph{FedAvg} is firstly proposed in \citep{mcmahan2017communication} as an aggregation algorithm for federated learning. The server updates the global model by a weighted average of the clients' model updates, and the weight value of each client is determined based on its data sample size. \cite{stich2018local} and \cite{woodworth2018graph} make the convergence analysis of \emph{FedAvg} on strongly-convex smooth loss functions, but they assume that the data are iid. The assumption has violated the characters of federated learning \citep{li2020federated}, \citep{kairouz2019advances}. And ~\cite{li2019convergence} makes the first convergence analysis of \emph{FedAvg} when the data are non-iid. But the ability of naive \emph{FedAvg} is very weak to resist Byzantine attacks.

In the iterative process of federated aggregation, honest clients send the true model updates to the server, wishing to train a global model by consolidating their private data. However, Byzantine clients attempt to perturb the optimization process \citep{chen2018internet}. Byzantine attacks may be caused by some data corruption events in the computing or communication process of the device, such as software crashes, hardware failures, and transmission errors. Simultaneously, they may also be caused by malicious clients through actively transmitting error information, thus misleading the learning process \citep{chen2018internet}. 

Byzantine-robust federated learning has received increasing attention in recent years. \emph{Krum} \citep{blanchard2017machine} is designed specially to defend Byzantine attacks in the federated learning. \emph{Krum} chooses one of the local model updates to generate the global model, the chosen local model update is closest to others. \emph{Krum} uses the Euclidean distance to determine which client updates should be chosen. \emph{GeoMed} \citep{chen2017distributed} uses the  geometric median which is a variant of the median from one dimension to multiple dimensions. Unlike the \emph{Krum}, the \emph{GeoMed} uses all client updates to generate a new global model which may not be one of the client updates. \emph{Trimmed Mean} \citep{yin2018byzantine} proposes that each dimension of its global model is obtained by averaging the parameters of clients' model updates in that dimension. But before calculating the average, the largest and smallest part of the parameters in that dimension are deleted, \cite{xie2018generalized} and \cite{mhamdi2018hidden} are all its variants. \emph{BREA} ~\citep{so2020byzantine} also considers the security of information transmission, but its defense method is still based on distance calculation.

All of the above defense methods based on statistical knowledge and distance are not effective in defending against Byzantine attacks in non-iid settings. \emph{Abnormal} \citep{li2019abnormal} uses an anomaly detection model to complete the detection of Byzantine attacks. In addition to the shortcomings in the training phase of the anomaly detection model, it has not given the reason and theoretical proof that the method can effectively defend against Byzantine attacks in non-iid setting.

The concept of independent and identically distributed (iid) of data is clear, but there are many meanings of non-iid. In this work, we only consider label distribution skew \citep{kairouz2019advances}. The categories of samples may vary across clients. For example, in the face recognition task, each user generally has their face data; for mobile device, some users may use emojis that do not show up in others' devices.
\par In this paper, we propose a method that combine \emph{credibility assessment} and \emph{unified update} to robust federated learning against Byzantine attacks on non-iid data. 

\section{Byzantine-robust Federated Learning on non-iid Data}    
We consider a federated setting with some clients and a server. For the rest of the paper, we will use the following  symbol definitions: $A$ is the total client set, $|A|=n$; $S$ is the selected client set in every iteration, $|S|=k$; among them, $B$ is Byzantine client set, $|B|=b$, and $H$ is honest client set, $|H|=h$. $w_i^t$ is the model update sent by the client $i$ to the server at round $t$, Byzantine attack rate $ \xi=\frac{b}{k}$. $W^t$ is the global model at round $t$, $D_p=\{D_1^p,...,D_n^p\}$ is clients' private data, $D_s=\{D_1^s,...,D_n^s\}$ is the clients' shared data, and data-sharing rate $\gamma=\frac{|D_s|}{|D_p|+|D_s|}$ ($|\centerdot|$ represents the sample size of the data set). 

\subsection{BRCA: Byzantine-robust Federated Learning via Credibility Assessment}
\begin{figure}[!h] 
\centering    
\includegraphics[width=6in]{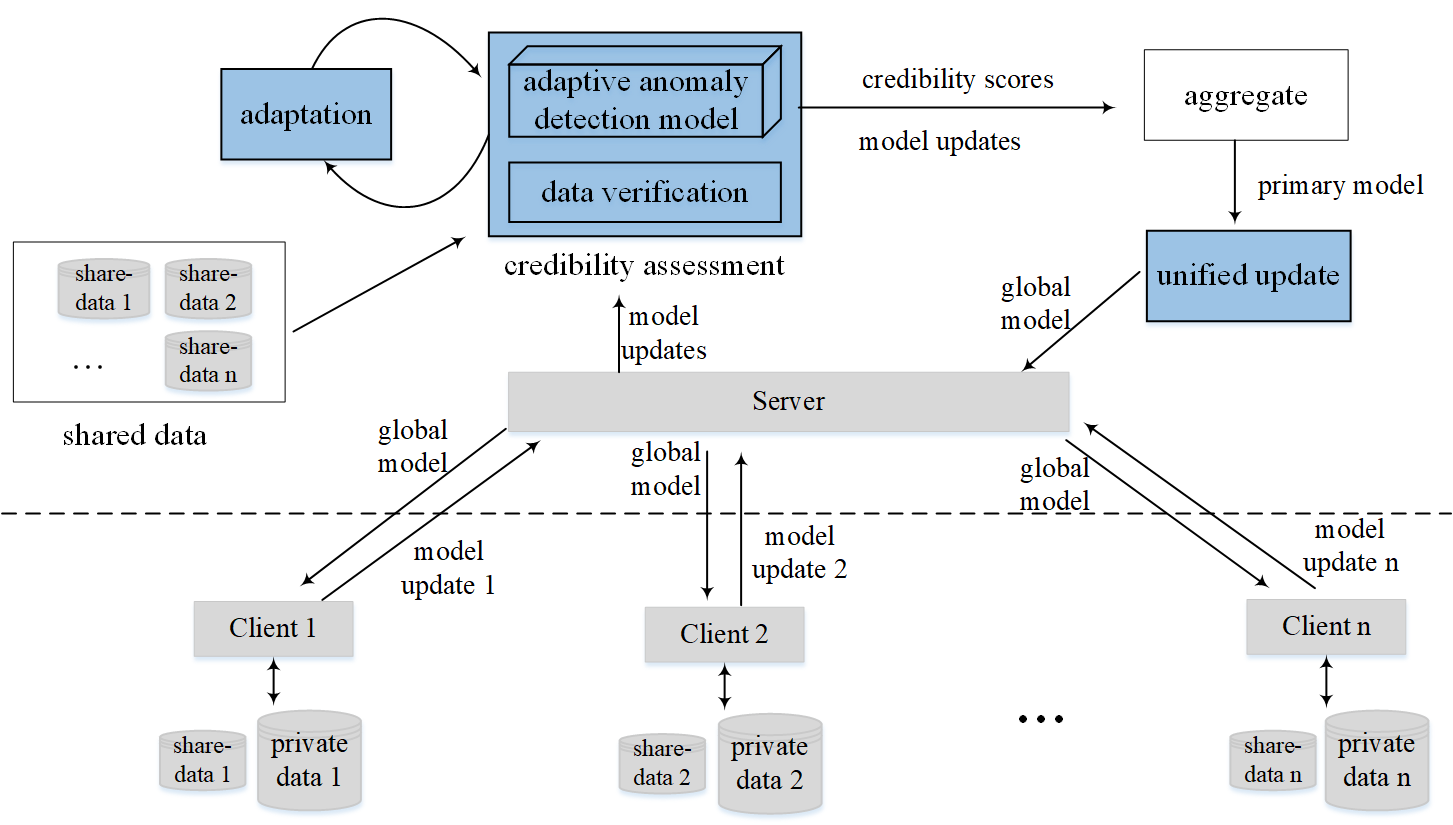}  
\caption{The frame diagram of the BRCA}     
\label{frame}     
\end{figure}

In order to enhance the robustness of  federated learning against Byzantines attacks on non-iid data, \emph{BRCA} combines \emph{credibility assessment} and \emph{unified update}, figure \ref{frame} is its frame diagram. 

Before training, each client needs to share some private data to the server. In each iteration, the server randomly selects some clients and sends the latest global model to them. These clients use their own private data to get the model updates on the global model, and send them to the server. After receiving model updates, the server conducts a \emph{credibility assessment} for each model update and calculates their credibility scores. Then use these credibility scores as the weights to aggregate the primary global model, considering momentum can effectively improve the ability of federated learning to resist Byzantine attacks~\citep{el2020distributed}. So our aggregation equation is as follow:
\begin{align}
    W^{t+1}=\alpha W^t+(1-\alpha)\sum_{i\in \boldsymbol{S} }r_i^t w_i^t \label{aggregation}
\end{align}
where $r_i^t$ is the credibility score of client $i$ at round $t$ and  $\alpha(0<\alpha<1)$ is a decay factor. Last, \emph{unified update} uses shared data to update the primary global model to get the new global model for this round.

\begin{algorithm2e}[!h]
\caption{BRCA }
\label{alg:alg1}
\LinesNumbered 
\KwIn{total clients $A$; total number of iterations $T$; learning rate $\eta_{server}$,$\eta_{client}$, $\eta_{detection}$; Byzantine attack rate $\xi$; epoch $E_{server}$, $E_{client}$; initial global model $W^0$; clients' private data $D_p=\left\{D_1^p,...,D_N^p\right\}$;
clients' shared data $D_s=\left\{D_1^s,...,D_n^s  \right\}$; initial anomaly detection model $\theta^0$; $\beta$; $\alpha$;$d$;$k$} 
\KwOut{global model $W^{T+1}$, anomaly detection model $\theta^{T+1}$ }
$R=\varnothing $ : the credibility score set.\\
$H=\varnothing $ : the honest client set.\\

\SetKwFunction{FMain}{AddAttack}
\SetKwProg{Fn}{Function}{:}{}
\Fn{\FMain{$w$}}{
    \textbf{return} $w$ attacked by Byzantine client.
}
\textbf{End Function} 

~\\
\For{each round t=0 to T}
{
     \underline{\textbf{Clients:}}    \\
     \For{client $i \in S$ \textbf{parallel}}{
          \For{each epoch $e=0$ to $E_{client}$}{
             $w_i^t= W^t-\eta_{client} \nabla \ell (D_i^p,W^t) $,where $\ell$ is the loss function.
          }
        \If {$i \in $ top $\xi $ percent of $S$}{
         $w_i^t= $ \FMain{$w_i^t$} 
         }
      \textbf{send} $w_i^t$ back to the server 
     \\
     }
     \underline{\textbf{Server:}}    \\
     \textbf{sample} $S \in A$ randomly  \\
     \textbf{broadcast} latest global model $W^t$ to each client $j \in S$  \\
     \textbf{receive} model updates from clients $Q=\{w_1^t,...,w_j^t...,w_k^t\}$, client $j \in S $ \\
     $R$, $H$, $\theta^{t+1}$ $=\emph{\textbf{Credibility Assessment}}\{ Q,D_s,\theta^t,\beta  , S, \eta_{detection} ,d\}$ \\
    $W^{t+1}=\alpha W^t+(1-\alpha)\sum_{client \  j\in S}r_j^t*w_j^t$, $r_j^t \in R$
    \\ 
    $W^{t+1} =\emph{\textbf{Unified Update}} \ (W^{t+1}, D_s, E_{server},\eta_{server}, H)$   \\
}

\Return global model $W^{T+1}$, anomaly detection model $\theta^{T+1}$.
\end{algorithm2e}

Algorithm \ref{alg:alg1} is the description of \emph{BRCA}, which contains \emph{Credibility Assessment} in line 22, and \emph{Unified Update} in line 24. The key of \emph{BRCA} to defend against Byzantine attacks is \emph{credibility assessment}. On non-iid data, the difference between client model updates is great, and it is difficult to judge whether the difference is caused by Byzantien attacks or the non-iid data. However, the model update of the honest client should have a good effect on its own private data, which is not affected by other clients. Simultaneously, anomaly detection model can effectively detect Byzantine attacks ~\citep{li2019abnormal}. So, we combine the above idea to detect Byzantine attacks. In order to solve the shortcomings of the existing anomaly detection models, we propose an adaptive anomaly detection model.

In this paper, the shared data is randomly selected by each client based on the sample category. Of course, other sampling methods could also be used, such as clustering. And, it must be pointed out that the shared data will only be used on the server, and other clients will not obtain it. That can protect the user's privacy to a certain extent.

\subsection{Credibility Assessment}
Algorithm \ref{alg:alg2}(\emph{Credibility Assessment}) is the key part of \emph{BRCA}, which assigns a credibility score for each client model update. A Byzantine client would be given much lower credibility score than an honest client. To ensure the accuracy of the credibility score, \emph{Credibility Assessment} integrates adaptive anomaly detection model and data verification. 

In algorithm \ref{alg:alg2}, line 4 is the \emph{data verification}, which calculates the verification score $f_i$ for the model update of each client $i$. And line 5 is the function \emph{get-anomaly-score()} of the adaptive anomaly detection model, which calculates detection score $e_i$. After this, the credibility $r_i$ of the model update is $r_i=\beta e_i + (1-\beta) f_i$, $R=\{r_1,...r_i...,r_k \}$, client $i \in S$. Function \emph{make-adaption()} in line 24 completes the adaption of the anomaly detection model.

In this paper, we judge the model update with a credibility score lower than the mean of $R$ as a Byzantine attack, and set its credibility score as zero. Finally, normalize the scores to get the final credibility scores.
 
\begin{algorithm2e}[!h]
\caption{Credibility Assessment}  
\label{alg:alg2}
\LinesNumbered 
\KwIn{local model updates $Q$; clients' shared data $D_s=\left\{D_1^s,...,D_n^s\right\}$; anomaly detection model $\theta^t$;
$\beta$; selected clients $S$; $\eta_{detection};d$ } 
\KwOut{credibility score of clients $R$; honest client set $H$; anomaly detection model $\theta^{t+1}$ }
     $R=\varnothing$: credibility score set;  $H= \varnothing$: the honest client set;  $sum=0$; $sum_e=0$;$sum_f=0$ \\
     $C=\{c_1^t,...,c_i^t,...,c_k^t\}$, client $i \in S$, $c_i^t$ is the weight of the last convolutional layer of $w_i^t$ \\
     
     \For{each client $i\in S$}{
     $\emph{\textbf{Data Verification}}$: \textbf{compute} $f_i^t$ with $w_i^t$ and $D_i^s$, client $i\in S$ base on equation \ref{eq:3} and equation \ref{eq:4}\\
     $e_i^t=\emph{\textbf{AADM.get-anomaly-score}} \ (\theta^t,C)$
     }
     $sum_e=\sum_{client \ i \in S } e_i^t $; $sum_f=\sum_{client \ i \in S } f_i^t$ \\
     
     \For{each client $i \in S$}{
     $e_i^t=e_i^t/ sum_e $ ; $f_i^t=f_i^t/ sum_f $ \\
     $r_i^t=\beta e_i^t + (1-\beta) f_i^t$ ; $R=R \cup \{r_i^t\}$
     }
     $M(R)$ is the mean of $R$   \\
     \For{each $r_i^t \in R$}{
     
     \uIf{$r_i^t \ < M(R)$}{
     $r_i^t=0$}
     \Else{
     $H=H \cup \{ i \}$
     }
     
     $sum+=r_i^t$
     }
     \For{each $r_i^t \in R$}{
     $r_i^t=r_i^t/ sum$
     }
    $\theta^{t+1}= \emph{\textbf{AADM.make-adaption}} \ (H,\theta^t,\eta_{detection},C,d) $ \\
\Return $R$, $H$,$\theta^{t+1}$.
\end{algorithm2e}
\subsubsection{Adaptive Anomaly Detection Model}
\par In the training process, we can not predict the type of attack, but we can estimate the model update of the honest client. So we can adopt a one-class classification algorithm to build the anomaly detection model using normal model updates. Such technique will learn the boundary of the model updates' distribution. Through this boundary, we can determine whether the new sample is abnormal. Autoencoder is known to be a one-class learning model that can effectively detect anomalies, especially for high-dimensional data~\citep{sakurada2014anomaly}.

In practical applications, we cannot get the target data to complete the pre-training of our anomaly detection model. Therefore, based on the idea of transfer learning, the initialized anomaly detection model will be pre-trained on the source data. Then, the server uses algorithm ~\ref{alg:alg3} to make the adaption of the anomaly detection model, which is implemented by fine-tuning on the target domain data. For these data involved in adaption, their credibility scores must be within a certain range. In the paper, we use the last convolutional layer of the model to complete the training and prediction of the anomaly detection model.

At round $t$, the detection score $e_i^t$ of client $i$  :
\begin{align}
    e_i^t=\Big( exp(\frac{ Mse(c_i^t - \theta^t(c_i^t))-\mu(E)}{\sigma(E)}) \Big)^{-2} 
    \label{eq: detection score}
\end{align}
where $\theta^t$ is the anomaly detection model, $c_i^t$ is the model update's last convolution layer, $Mse(c_i^t - \theta^t(c_i^t))$ represents reconstruction error. $E=\left\{e_1^t,...,e_i^t,...,e_k^t \right\}$, client $i \in S$, $\mu(E)$,$\sigma(E)$ are the mean and variance of set $E$. 

\begin{algorithm2e}[!h]
\caption{AADM Adaptive Anomaly Detection Model}  
\label{alg:alg3}
\LinesNumbered 
\KwIn{anomaly detection model $\theta^t$; weights of the last convolutional layer of the local model $C$ ; $\eta_{detection}$; credibility score $R$; honest client set $H$;$d$ } 
\KwOut{updated anomaly detection model $\theta^{t+1}$ }
    
\SetKwFunction{FMain}{get-anomaly-score}
\SetKwProg{Fn}{Function}{:}{}
\Fn{\FMain{$\theta^t,c_i^t$}}{
    \textbf{compute} $e_i^t$ with $\theta^t$ and $c_i^t$, client $i \in S$ based on equation \ref{eq: detection score} \\
    \Return $e_i^t$.
}
\textbf{End Function} 

~\\

\SetKwFunction{FMain}{make-adaption}
\SetKwProg{Fn}{Function}{:}{}
\Fn{\FMain{$H,\theta^t,\eta_{detection},C,d$}}{
    $H_o$ is a subset of $H$ obtained by removing the
the clients whose credibility score is in the largest and smallest $d$ fraction. \\
       \For{$client \ i \in H_o$}{
    $\theta^{t+1}=\theta^t-\eta_{detection} \nabla \ell (\theta^t,c_i^t)$ }
     \Return updated anomaly detection model $\theta^{t+1}$.
}
\textbf{End Function} 
\end{algorithm2e}
Our anomaly detection model is different from the one in \emph{Abnormal}:
(1) \emph{Abnormal} uses the test set of the data set to train the anomaly detection model. Although the detection model obtained can complete the detection task very well, in most cases the test data set is not available. Therefore, based on the idea of transfer learning, we complete the pre-training of the anomaly detection model in the source domain.
(2) \emph{Abnormal}'s anomaly detection model will not be updated after training on the test set. We think this is unreasonable, because the test set is only a tiny part of the overall data. Use a small part of the training data to detect most of the remaining data, and the result may not be accurate enough. Therefore, first, pre-training of the anomaly detection model is completed in the source domain. Second we use the data of the target domain to fine-tune it in the iterative process to update the anomaly detection model dynamically, as shown in function \emph{make-adaption} in Algorithm~\ref{alg:alg3}. 

\subsubsection{Data Verification}
\par The non-iid of client data increases the difficulty of Byzantine defense. But the performance of the updated model of each client on its own shared data is not affected by other clients, which can be effectively solved this problem. Therefore, we use the clients' shared data $D_{s}=\left\{D_1^s,...,D_i^s,...,D_k^s\right\}$, client $i\in S$ to calculate the verification score of their updated model:
\begin{align}
    f_i^t=\Big( exp(\frac{l_i^t-\mu(L)}{\sigma(L)}) \Big)^{-2}
    \label{eq:3}
\end{align}
where $l_i^t$ is loss of client $i$ calculated on model $w_i^t$ using the shared data $D_i^s$ at round $t$: 
\begin{align}
    l_i^t=\frac{1}{| D_i^s |} \sum_{j=0}^{|D_i^s|}
    \ell(D_i^{s(j)},w_i^t)
    \label{eq:4}
\end{align}
where $D_i^{s(j)}$ is the $j$-th sample of $D_i^s$ and $\mu(L)$,$\sigma(L)$ are the mean and variance of set $L=\{l_1,...,l_k\}$ respectively.

\subsection{Unified Update}
\par After getting the credibility score $r_t^k$ in algrithm~\ref{alg:alg2} with the anomaly score $e_t^k$ and the verification score $f_t^k$, we can complete the aggregation of the clients' local model updates in equation ~\ref{aggregation} and get a preliminary updated global model. However, due to the non-iid of client data, the knowledge learned by the local model of each client has certain limitations, and the model differences between two clients are also significant. Therefore, to solve the problem that the preliminary aggregation model lacks a clear and consistent goal, we introduce an additional \emph{unified update} procedure with shared data on server, details can be seen in algorithm \ref{alg:alg4}.

Because the data used for the \emph{unified update} is composed of each client's data, it can more comprehensively cover the distribution of the overall data. The goal and direction of the \emph{unified update} are based on the overall situation and will not tend to be in an individual area.

\begin{algorithm2e}
\caption{Unified Update} 
\label{alg:alg4}
\LinesNumbered 
\KwIn{global model $W^{t+1}$; clients' shared data $D_s=\left\{D_1^s,...,D_n^s  \right\}$; $E_{server}$;$\eta_{server}$; honest client set $H$ } 
\KwOut{global model $W^{t+1}$.}
    \For{each epoch $e=0$ to $E_{server}$}{
          
          \For{$i \in H$ }{
             $W^{t+1}= W^{t+1}-\eta_{server} \nabla \ell (D_i^s,W^{t+1}) $
          }
     } 
\Return global model $W^{t+1}$.
\end{algorithm2e}

\section{Experiments}
To verify the effectiveness of \emph{BRCA}, we structure the client's data into varying degrees of non-iid. At the same time, we also explored the impact of different amounts of shared data on the global model and compared the performance of our anomaly detection model with the \emph{Abnormal}'s.
\subsection{Experimental steup}
\subsubsection{Datasets}
We do the experiments on Mnist and Cifar10, and construct four different data distributions: (a) non-iid-1: each client only has one class of data. (b) non-iid-2: each client has 2 classes of data. (c) non-iid-3: each client has 5 classes of data. (d) iid: each client has 10 classes of data. 

For Mnist, 100 clients are set and four data distributions are: (a) non-iid-1: each class of data in the training dataset is divided into 10 pieces, and each client selects one piece as its private data. (b) non-iid-2: each class of data in the training dataset is divided into 20 pieces, and each client selects 2 pieces of different classes of the data. (c) non-iid-3 each class of data in the training dataset is divided into 50 pieces, and each client selects 5 pieces of different classes of the data. (d) iid: each class of data in the training dataset is divided into 100 pieces, and each client selects 10 pieces of different classes of the data. As for the source domains used for the pre-training of the anomaly detection model, we randomly select 20,000 lowercase letters in the Nist dataset. 

For Cifar10, there are 10 clients and the construction of four data distributions is similar to that of the Mnist. We select some classes of data in Cifar100 as source domain, which are as follows: lamp (number:40), lawn mower (41), lobster (45), man (46), forest (47), mountain (49), girl (35) , Snake (78), Rose (70) and Tao (68), these samples do not appear in Cifar10.

\subsubsection{Models}
\par We use logistic regression on Mnist dataset. $\eta_{server}=0.1$, $\eta_{client}=0.1$, $\eta_{detection}=0.02$, $E_{client}=5$, $E_{server}=1 $,$n=100$, $k=30$, $\xi=20\%$. Two convolution layers and three fully connected layer on Cifar10, $\eta_{server}=0.05$, $\eta_{client}=0.05$, $\eta_{detection}=0.002$, $E_{client}=10$, $E_{server}=1$, $n=10$, $k=10$, $\xi=20\%$. The structure of models are the same as \citep{li2019convergence}.

\subsubsection{Benchmark Byzantine attacks}
\par \emph{Same-value attacks}: A Byzantine client $i$ sends the model update $w_i=c\boldsymbol{1}$ to the server($\boldsymbol{1}$ is all-one vectors, c is a constant), we set $c=5$. \emph{Sign-flipping attacks}: In this scenario, each client $i$ computes its true model update $w_i$, then Byzantine clients send $w_i=a w_i(a<0)$ to the server, we set $a=-5$. \emph{Gaussian attacks}: Byzantine clients add Gaussian noise to all the dimensions of the model update $w_i=w_i+\epsilon$, where $\epsilon$ follows Gaussian distribution $~N(0,g^2)$ where $g$ is the variance, we set g=0.3.

\subsubsection{Benchmark defense methods}
\par Defenses: \emph{Krum}, \emph{GeoMed}, \emph{Trimmed Mean},  \emph{Abnormal} and \emph{No Defense}. \emph{No Defense} does not use any defense methods. 

\subsection{Result and Discussion}
\subsubsection{\textbf{Impact of shared data rate}}
\par
\begin{figure}[!h] \centering    
\subfigure[Same value] {
 \label{fig:a}     
\includegraphics[width=1.8in]{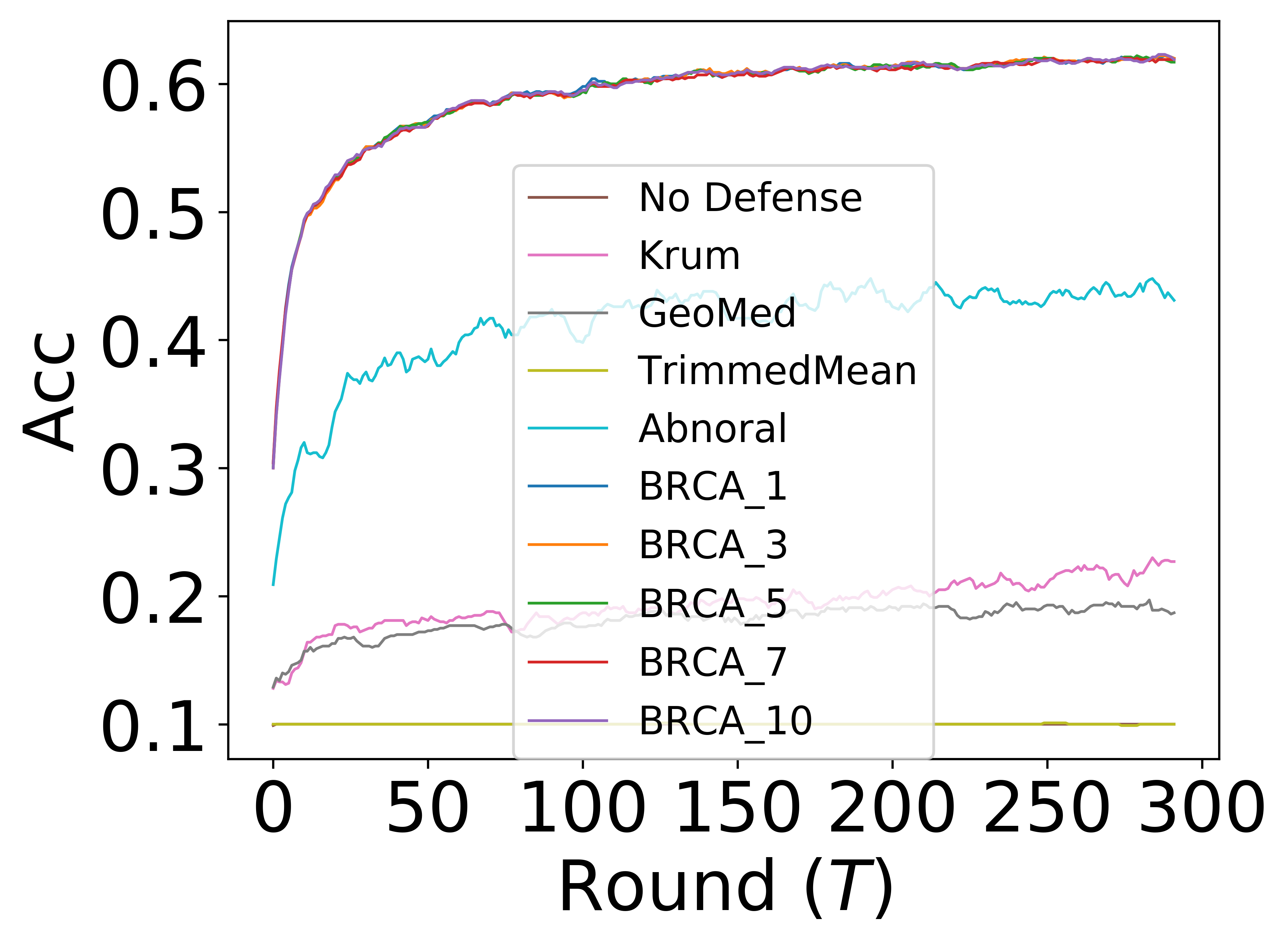}  
}     
\subfigure[Sign flipping] { 
\label{fig:b}     
\includegraphics[width=1.8in]{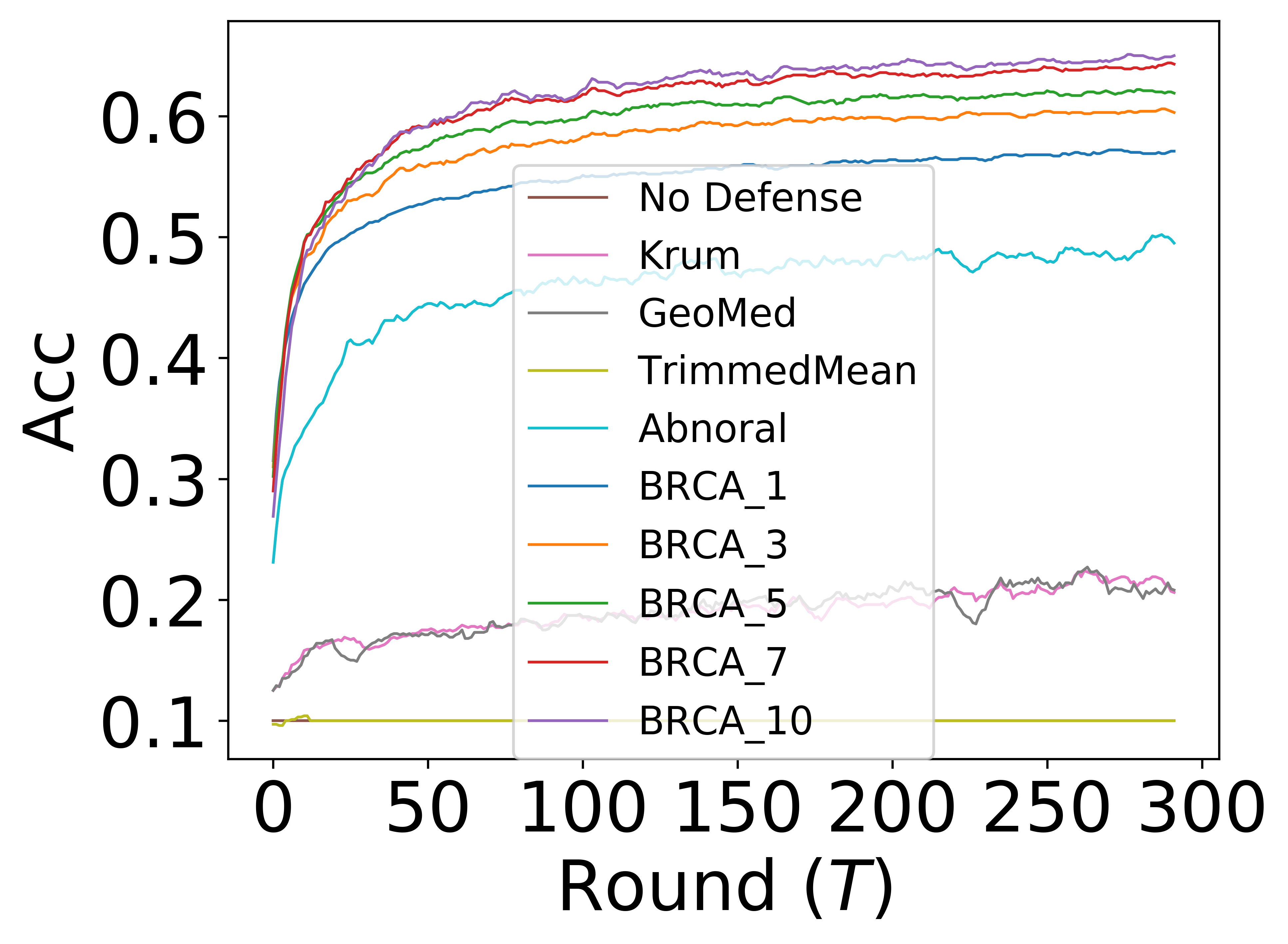}     
}    
\subfigure[Gaussian noisy] { 
\label{fig:c}     
\includegraphics[width=1.8in]{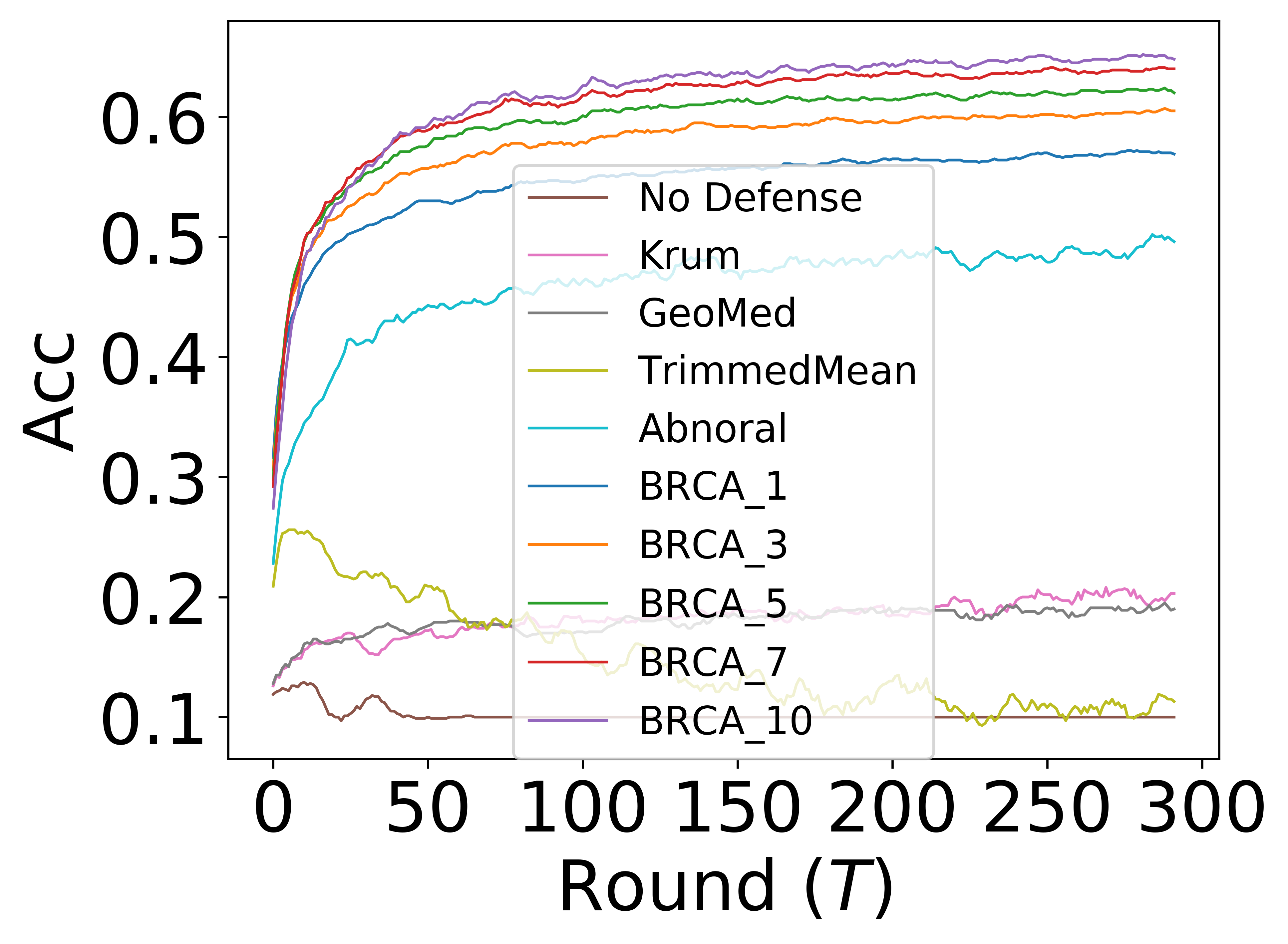}     
}   
\caption{The Accuracy of Cifar10. Byzantine attack types from $(a)$ to $(c)$ are as follows: Same value, Sign flipping and Gaussian noisy. Six defense methods are adopted for each type of attack, in order: No defense, Krum, GeoMed, Trimmed Mean, abnormal, Abnormal and BRCA. For
Ours, there are five different shared data rate (1\%; 3\%; 5\%; 7\%; 10\%), which correspond accordingly: BRCA\_1, BRCA\_3, BRCA\_5, BRCA\_7, BRCA\_10. Figure 3 is the same, so we will not explain it.}     
\label{acc of cifar10}     
\end{figure}

\begin{figure}[!h] \centering    
\subfigure[Same value] {
 \label{fig:a}     
\includegraphics[width=1.8in]{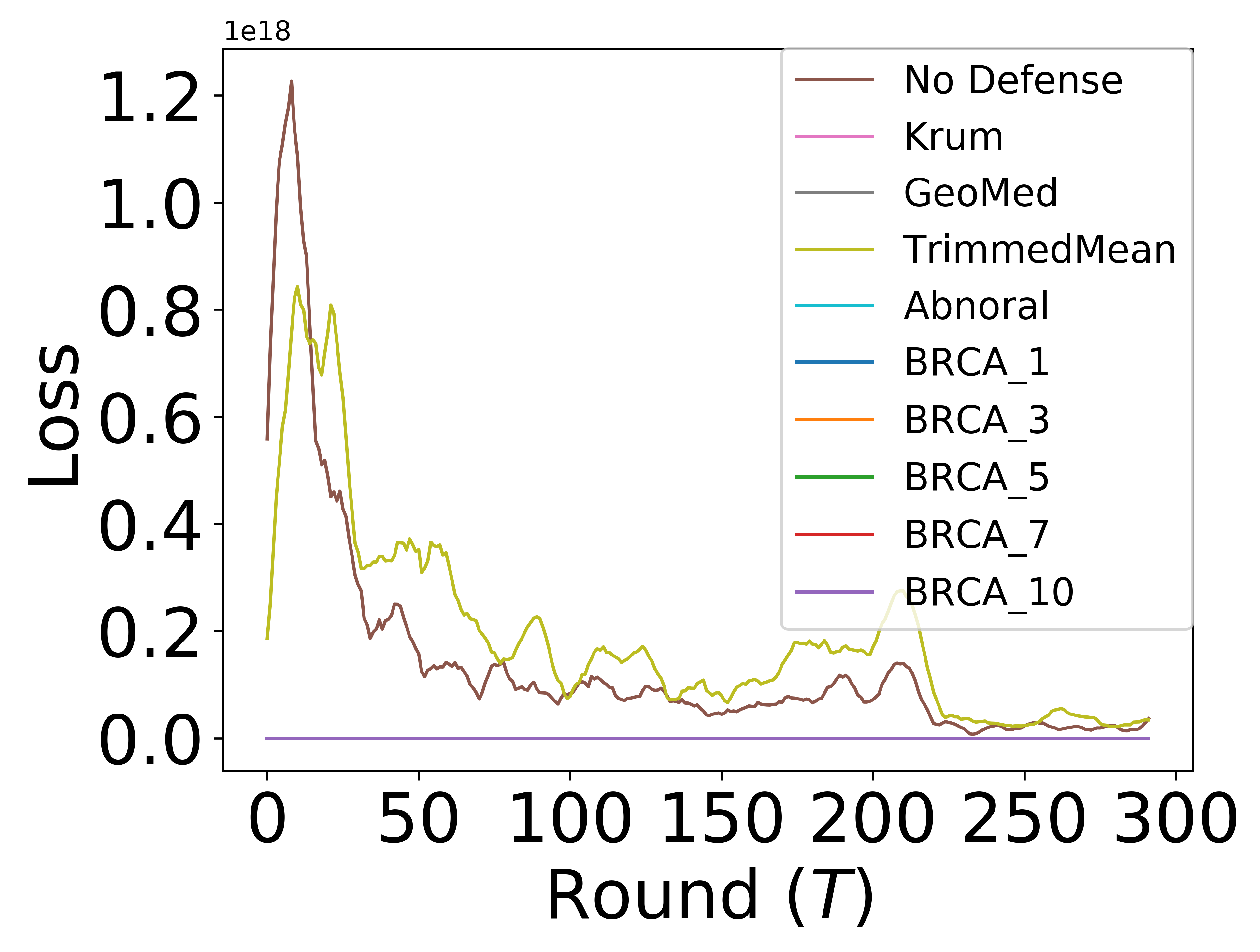}  
}     
\subfigure[Sign flipping] { 
\label{fig:b}     
\includegraphics[width=1.8in]{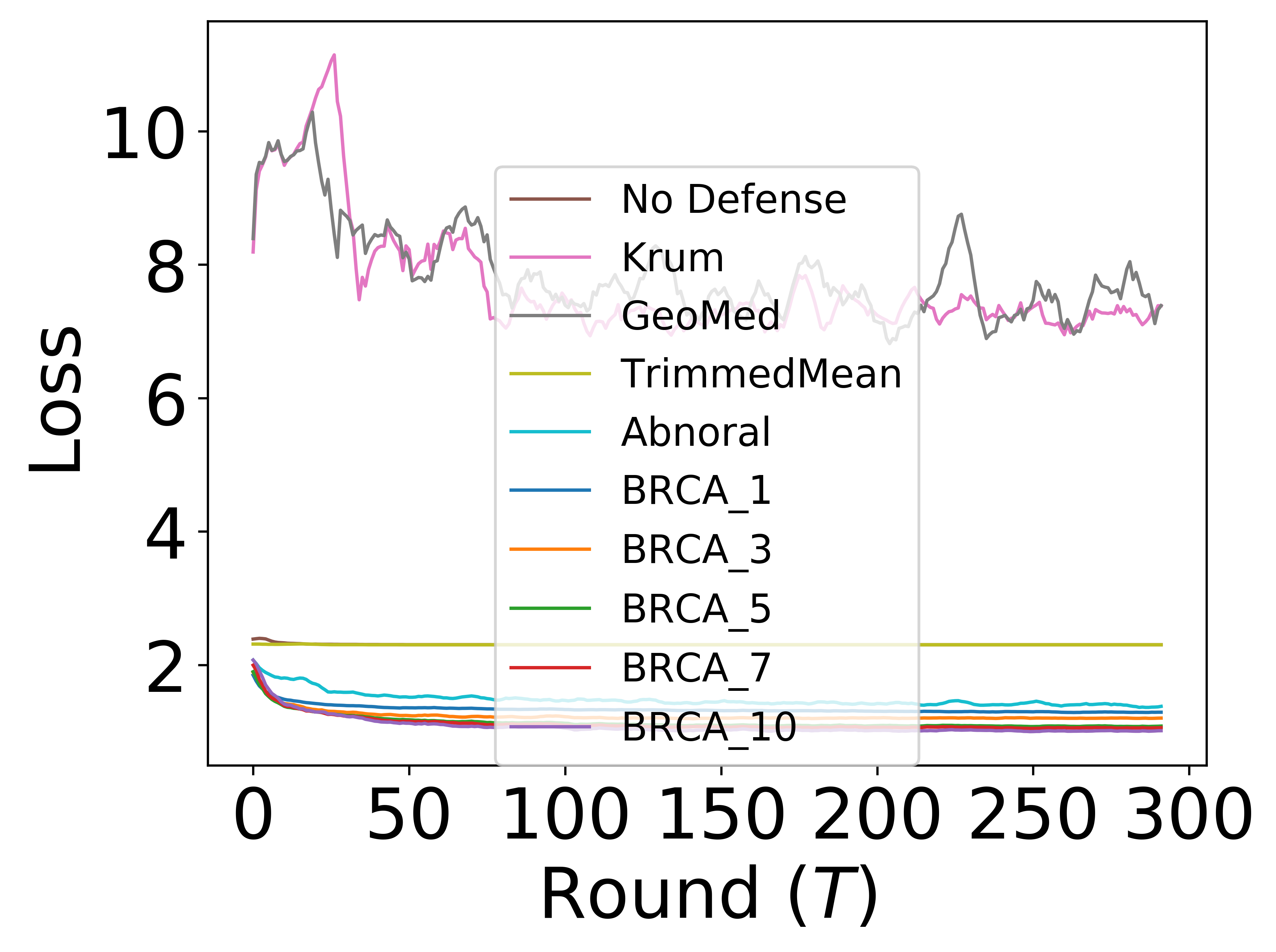}     
}    
\subfigure[Gaussian noisy] { 
\label{fig:c}     
\includegraphics[width=1.8in]{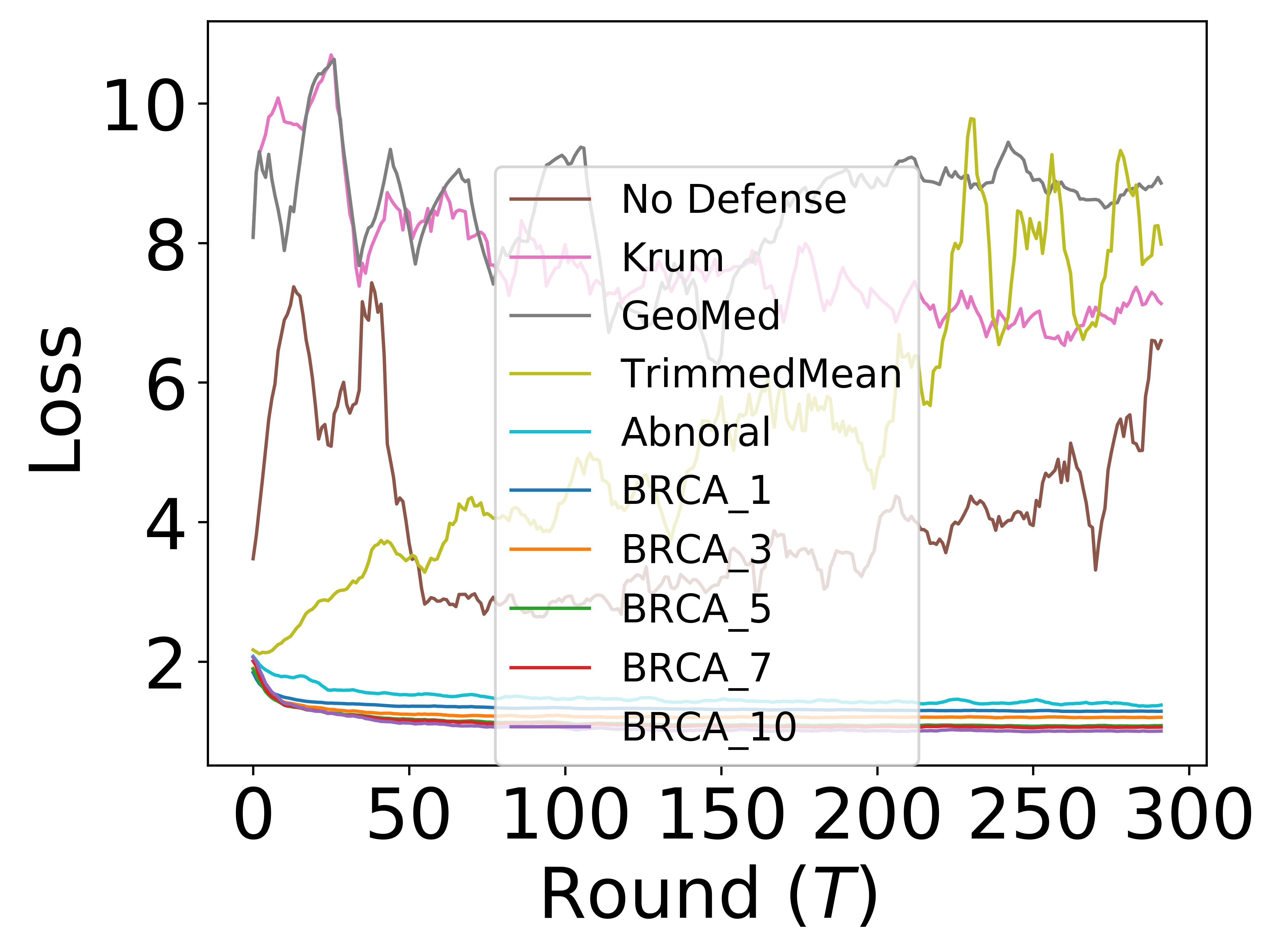}     
}

\caption{The loss of Cifar10.  }     
\label{loss of cifar10}     
\end{figure}

 In the first experiment, we test the influence of the shared data rate $\gamma$ in our algorithm, and it is performed under the data distribution of non-iid-2. We implement it on five different values [1\%,3\%,5\%,7\%,10\%]. Figure \ref{acc of cifar10} and figure \ref{loss of cifar10} are the accuracy and loss for Cifar10. 
It is found that: 
1) In all cases of Byzantine attacks, our algorithm is superior to the three benchmark methods. 
2) Only 1\% of the data shared by the client can significantly improve the performance of the global model. For three Byzantine attacks, \emph{Krum}, \emph{GeoMed}, \emph{Trimmed Mean}, \emph{No Defense} are all unable to converge. This also shows that when the model is complex, such methods to defend against Byzantine attacks will be significantly reduced.
3) With the increase in the client data sharing ratio, the performance of the global model has become lower. When the client shares the data ratio from $1\%$ to $10\%$, the average growth rate under the three Byzantine attacks are: $1.8\% \rightarrow 1.41\% \rightarrow 0.97\% \rightarrow 0.92\%$. The clients only share one percent of the data, and the performance of the global model can be greatly improved. 

\begin{figure}[!h] \centering    
\subfigure[Same value] {
 \label{fig:a}     
\includegraphics[width=1.8in]{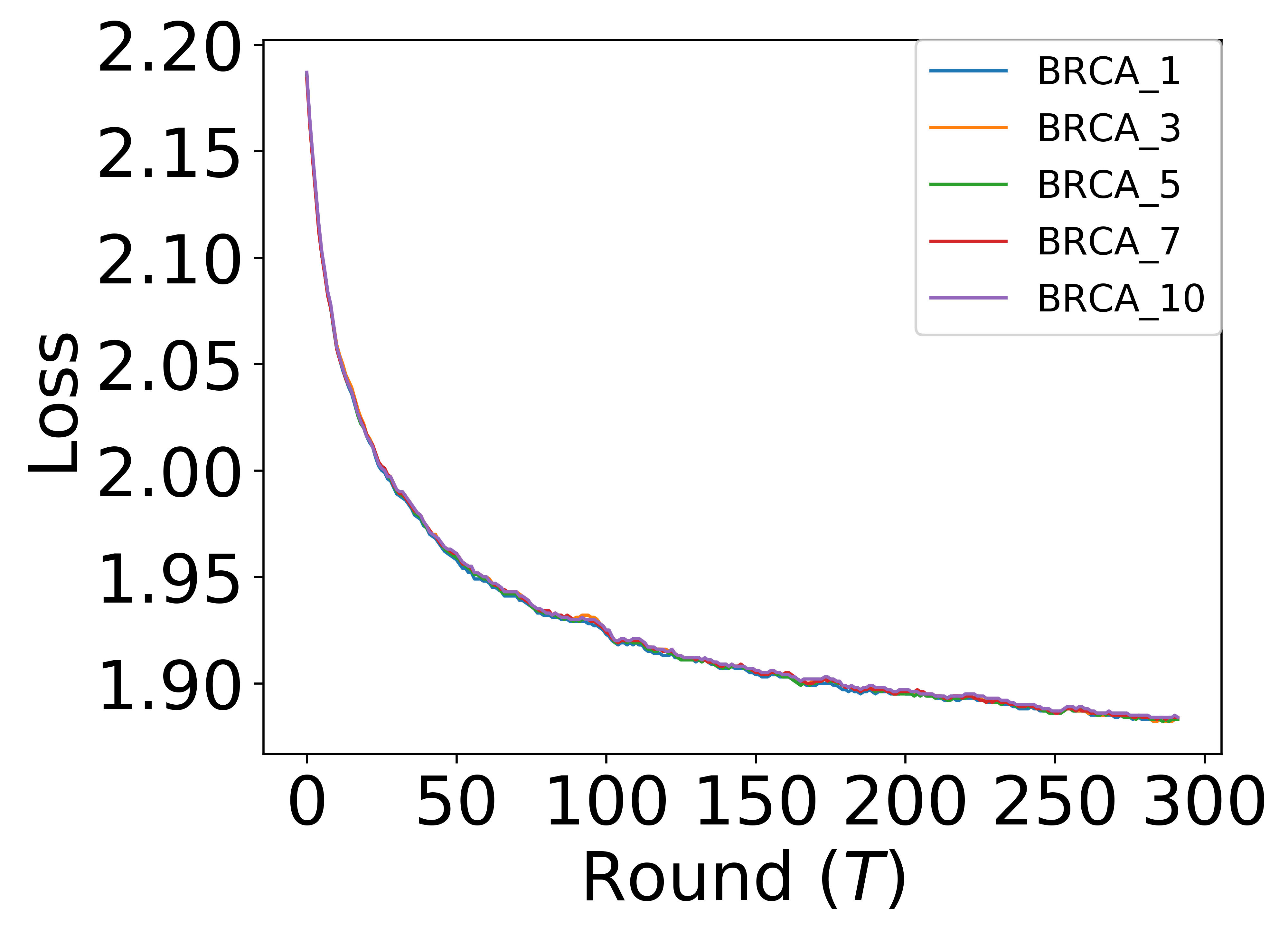}  
}     
\subfigure[Sign flipping] { 
\label{fig:b}     
\includegraphics[width=1.8in]{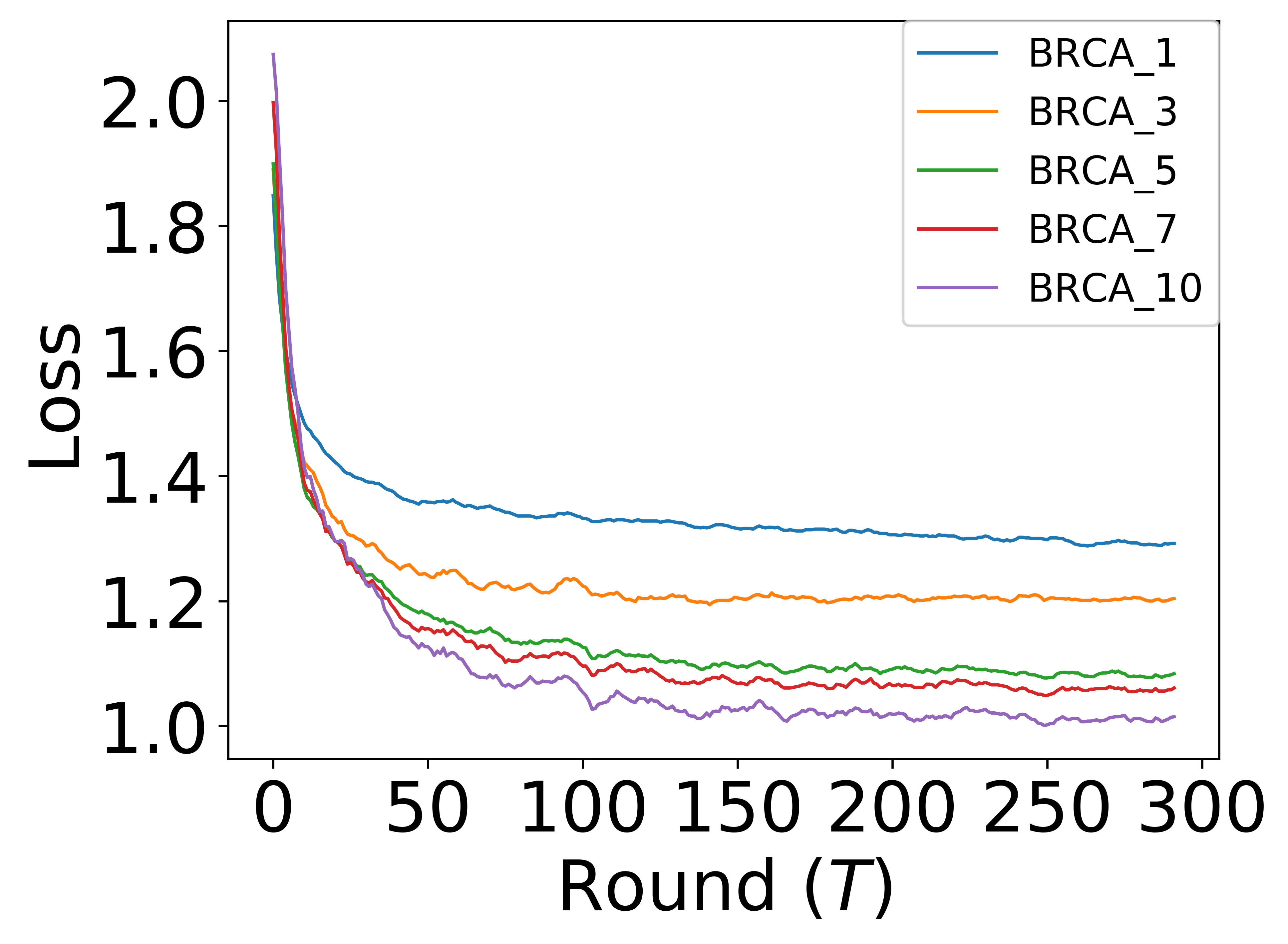}     
}    
\subfigure[Gaussian noisy] { 
\label{fig:c}     
\includegraphics[width=1.8in]{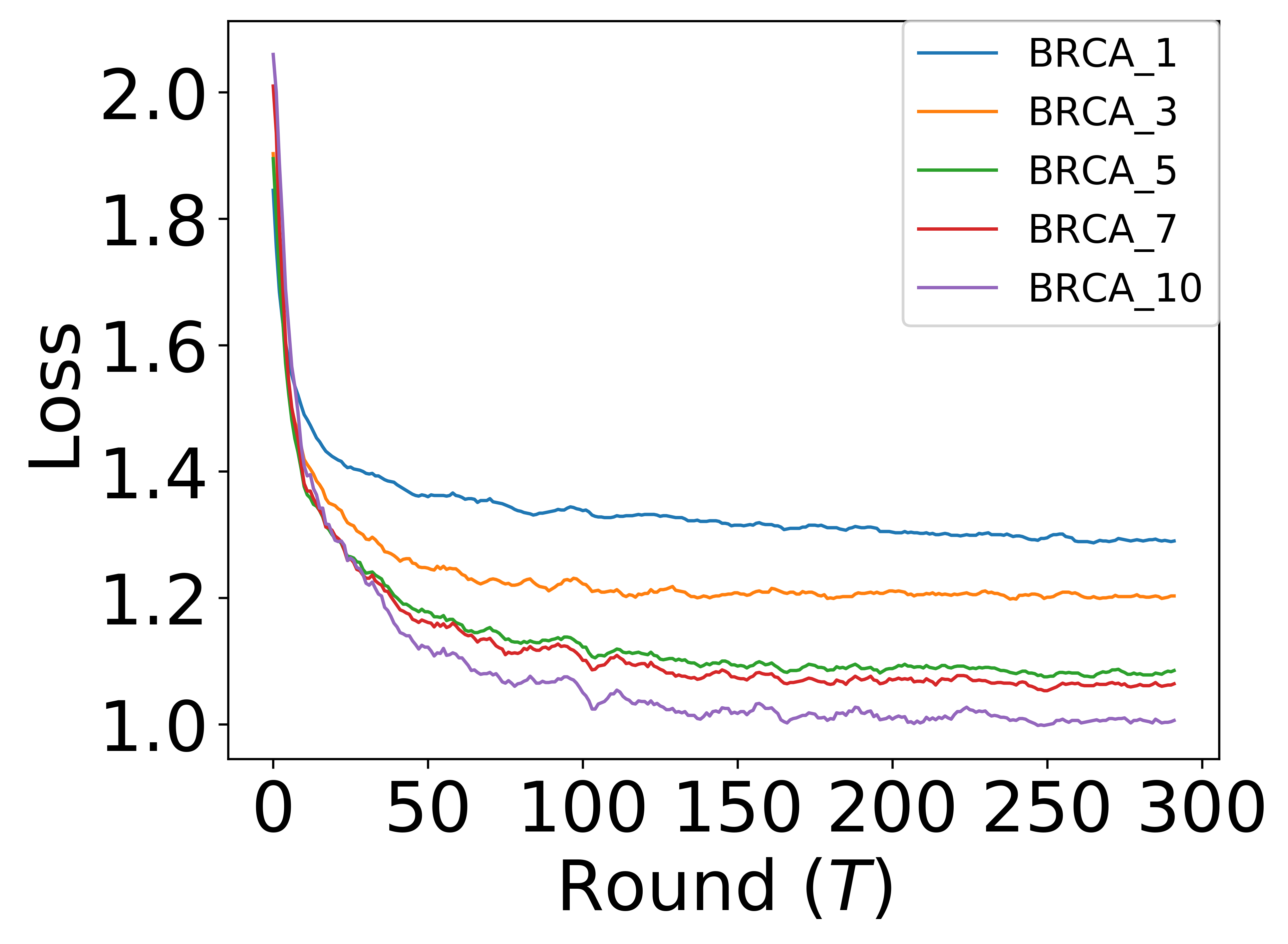}     
}

\caption{The loss of BRCA on Cifar10 with five different shared rate.  }     
\label{loss of cifar10(ours)}     
\end{figure}

Figure \ref{loss of cifar10(ours)} more clearly shows the impact of different shared data rate on the loss value of the global model on Cifar10. 

\subsubsection{\textbf{Performance of anomaly detection model}}

In this part, the purpose of our experiment has two: 1) Compare anomaly detection model between ours and \emph{Abnormal}.  2) Explore the robustness of the anomaly detection model to data that are non-iid.

\par In order to objectively and accurately compare the detection performance of the anomaly detection model against Byzantine attacks between \emph{BRCA} and \emph{Abnormal}, we use the cross-entropy loss as the evaluation indicator which is only calculated by the detection score. 
Firstly, we get detection scores $E=\{e_1,...,e_i,...,e_k \}$ based on
model update $w_i$ and $\theta$, client $i \in S$. Then, we set $P=Sigmoid(E-\mu(E))$ represents the probability that the client is honest and $1-P$ is the probability that the client is Byzantine. Lasty, we use $P$ and true label $Y$ $(y^i=0,i\in B$ and $y^j=1,j \in H)$ to calculate the cross entropy loss $l=\sum_{i=1}^{k} y_i ln(p_i)$.

\par
\begin{figure}[H] \centering
\subfigure[Same value] { 
\label{fig:a}     
\includegraphics[width=1.8in]{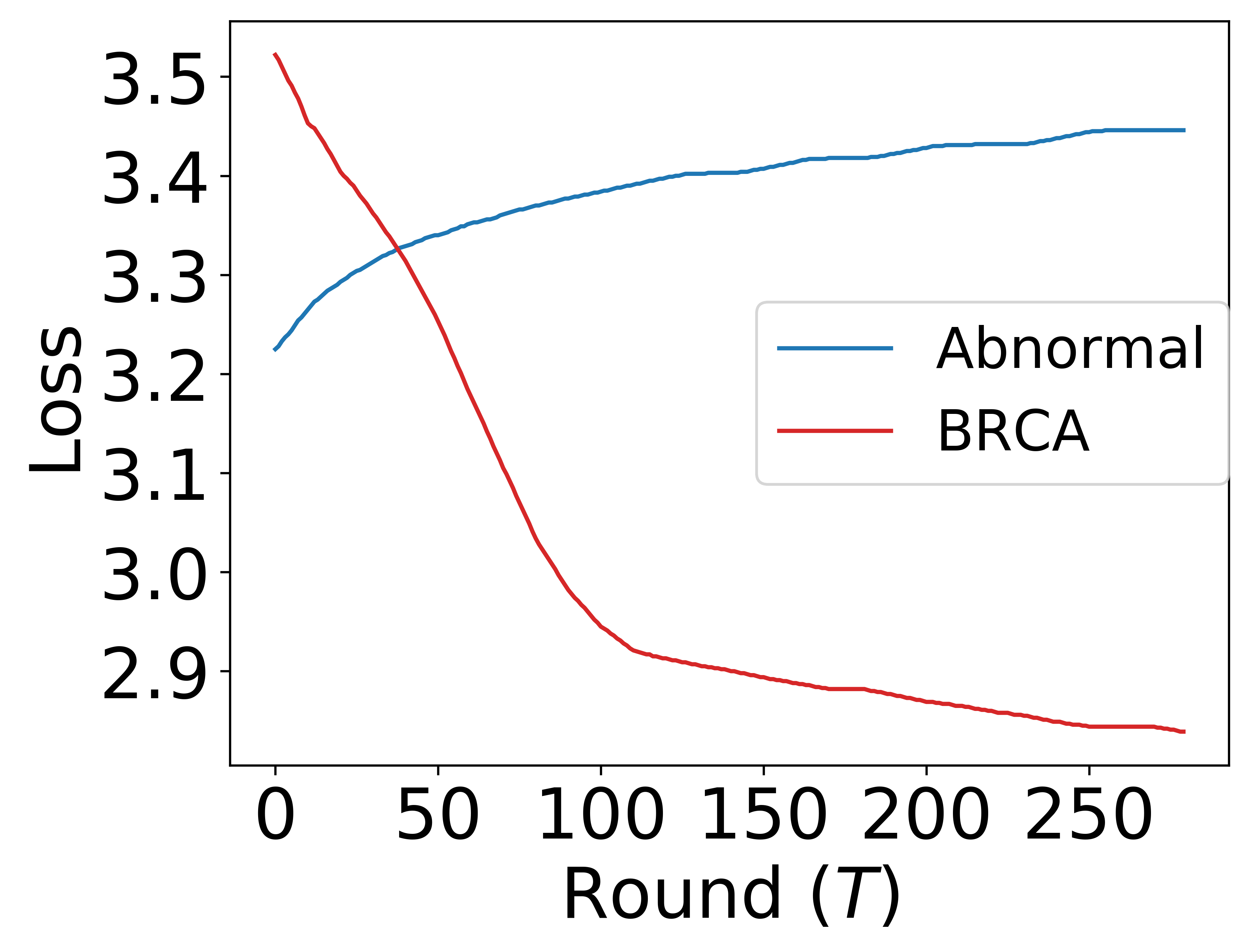}   
}  
\subfigure[Sign flipping] { 
\label{fig:b}     
\includegraphics[width=1.8in]{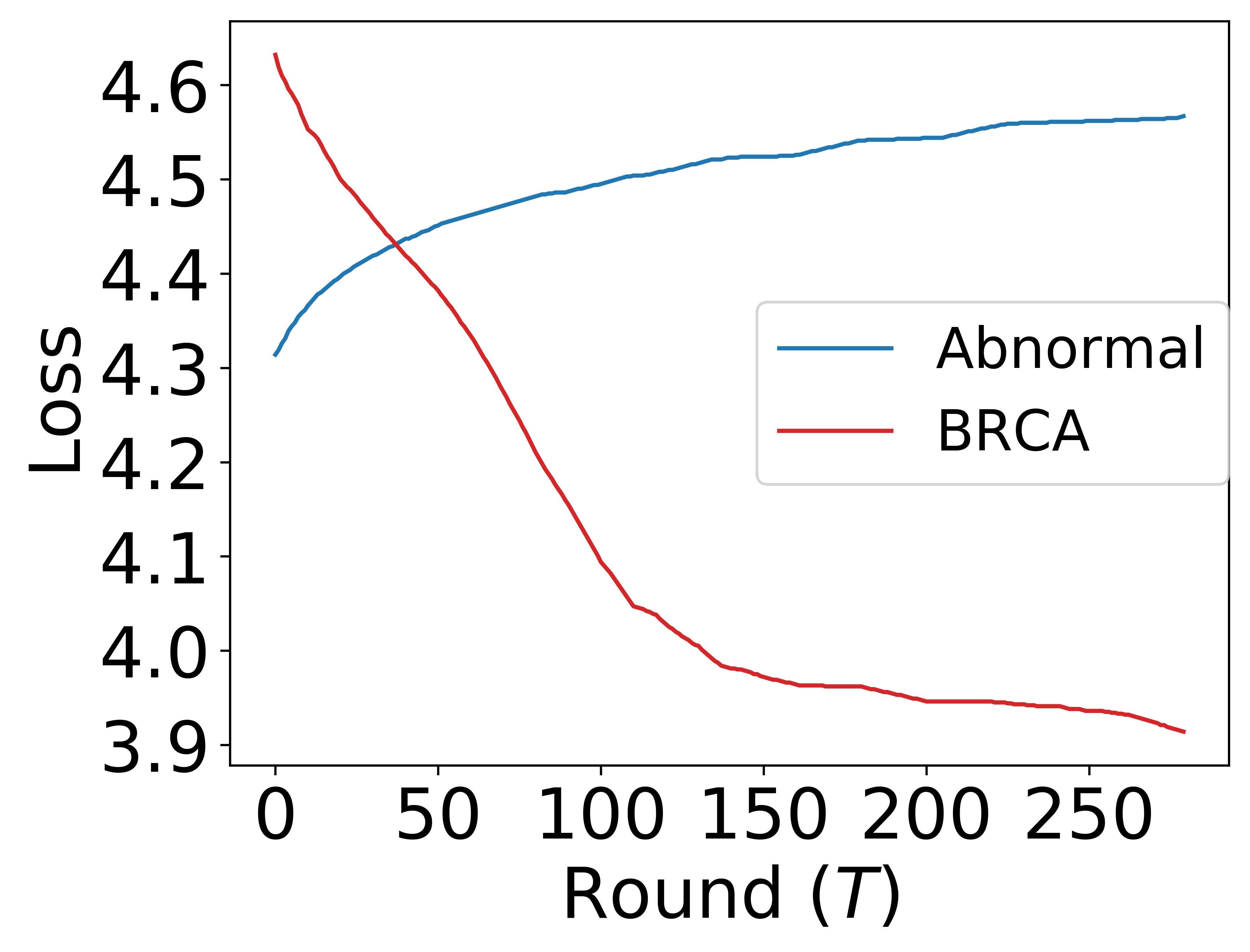}
}  
\subfigure[Gaussian noisy] { 
\label{fig:b}     
\includegraphics[width=1.8in]{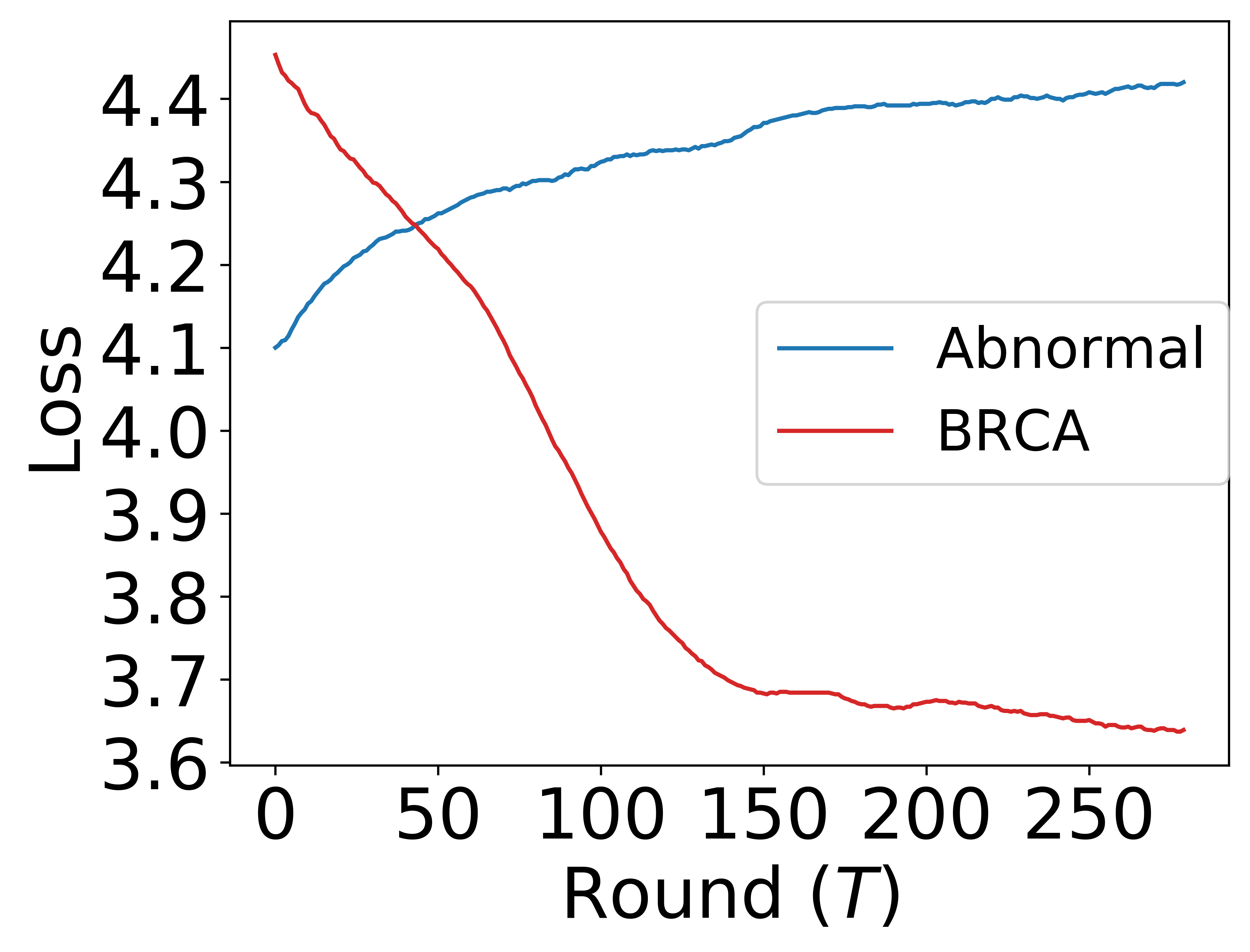}
}  

\caption{the cross entropy loss of our and \emph{Abnormal} anomaly detection model, on Cifar10 with non-iid-2. $(a)$, $(b)$, $(c)$ are the performance for three Byzantine attacks(Gaussian noisy, sign flippig and same value) }     
\label{ours Vs abnormal non-iid2}     
\end{figure}

 Figure \ref{ours Vs abnormal non-iid2} $(a)$, $(b)$ and $(c)$ compare the loss of the anomaly detection model between \emph{BRCA} and the \emph{Abnormal}. From the figures, we can see that our model has a greater loss than Abnormal in the initial stage, mainly due to the pre-training of the anomaly detection model using the transfer learning. The initial pre-trained anomaly detection model cannot be used well in the target domain. But with the adaptation, the loss of our model continues to decrease and gradually outperforms the \emph{Abnormal}. Although \emph{Abnormal} has a low loss in the initial stage, as the training progresses, the loss gradually increases, and the detection ability continues to decrease. 

\begin{figure}[!h] \centering
\subfigure[Same value] { 
\label{fig:c}     
\includegraphics[width=1.8in]{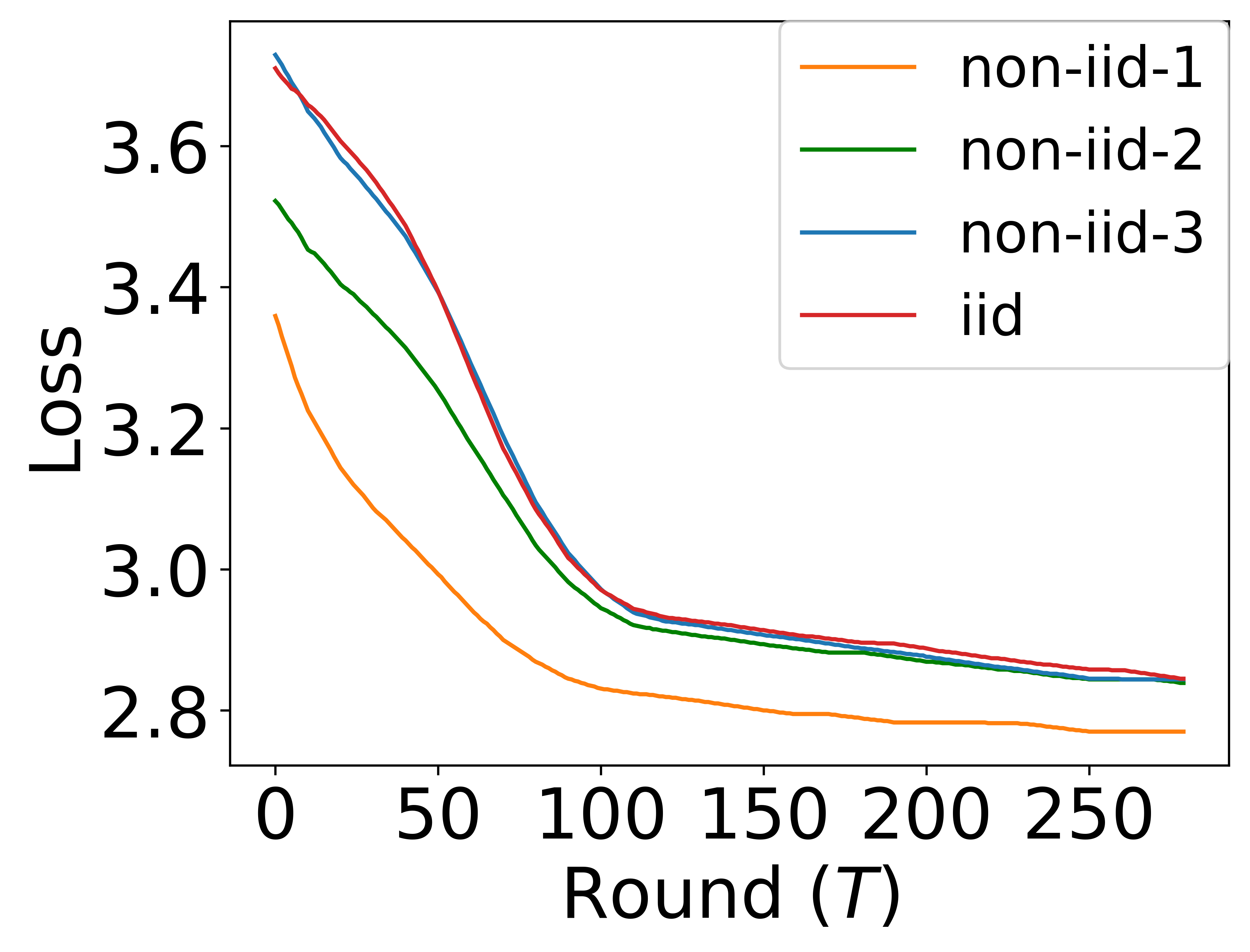}     
}    
\subfigure[Sign flipping] { 
\label{fig:b}     
\includegraphics[width=1.8in]{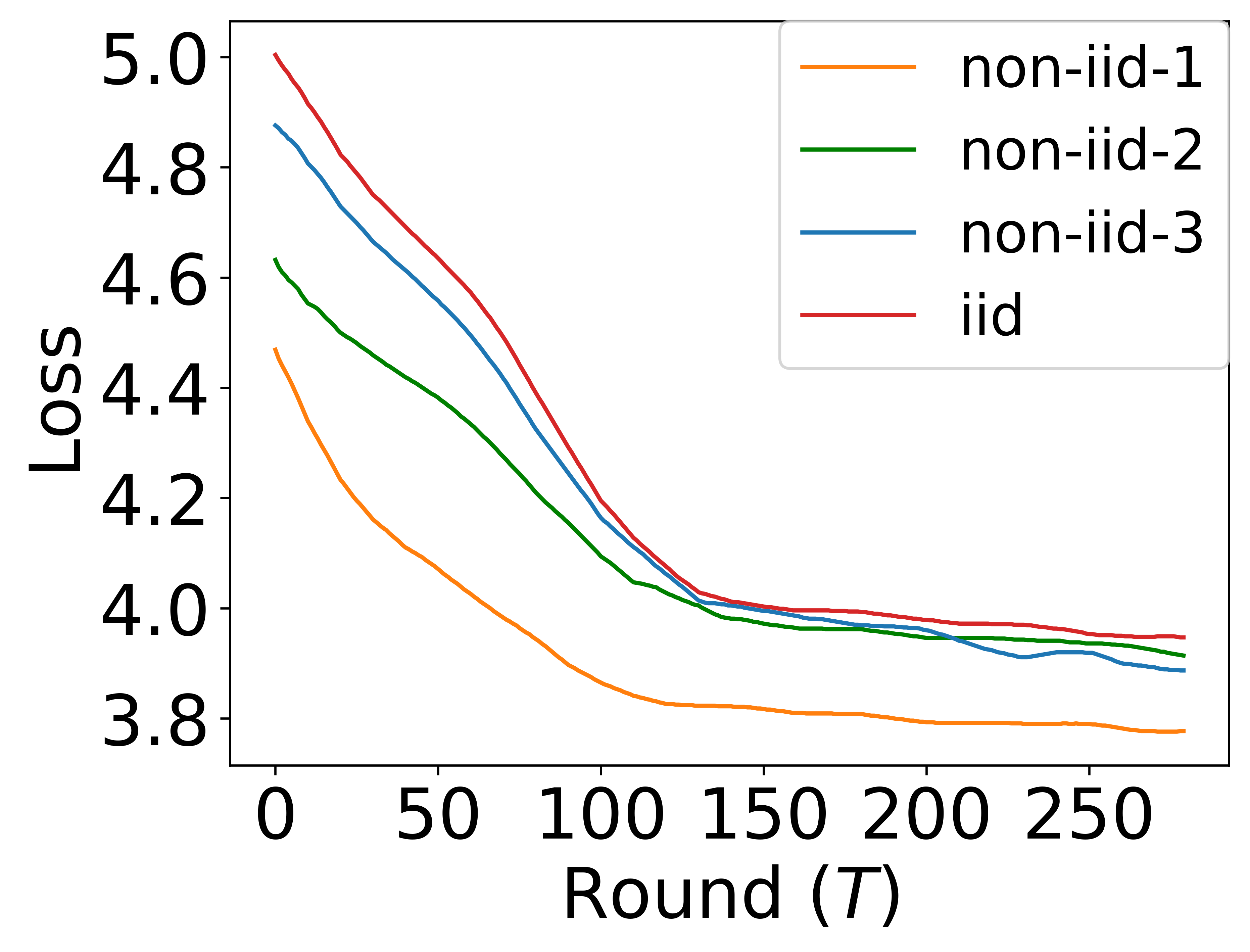}     
}    
\subfigure[Gaussian noisy] {
 \label{fig:a}     
\includegraphics[width=1.8in]{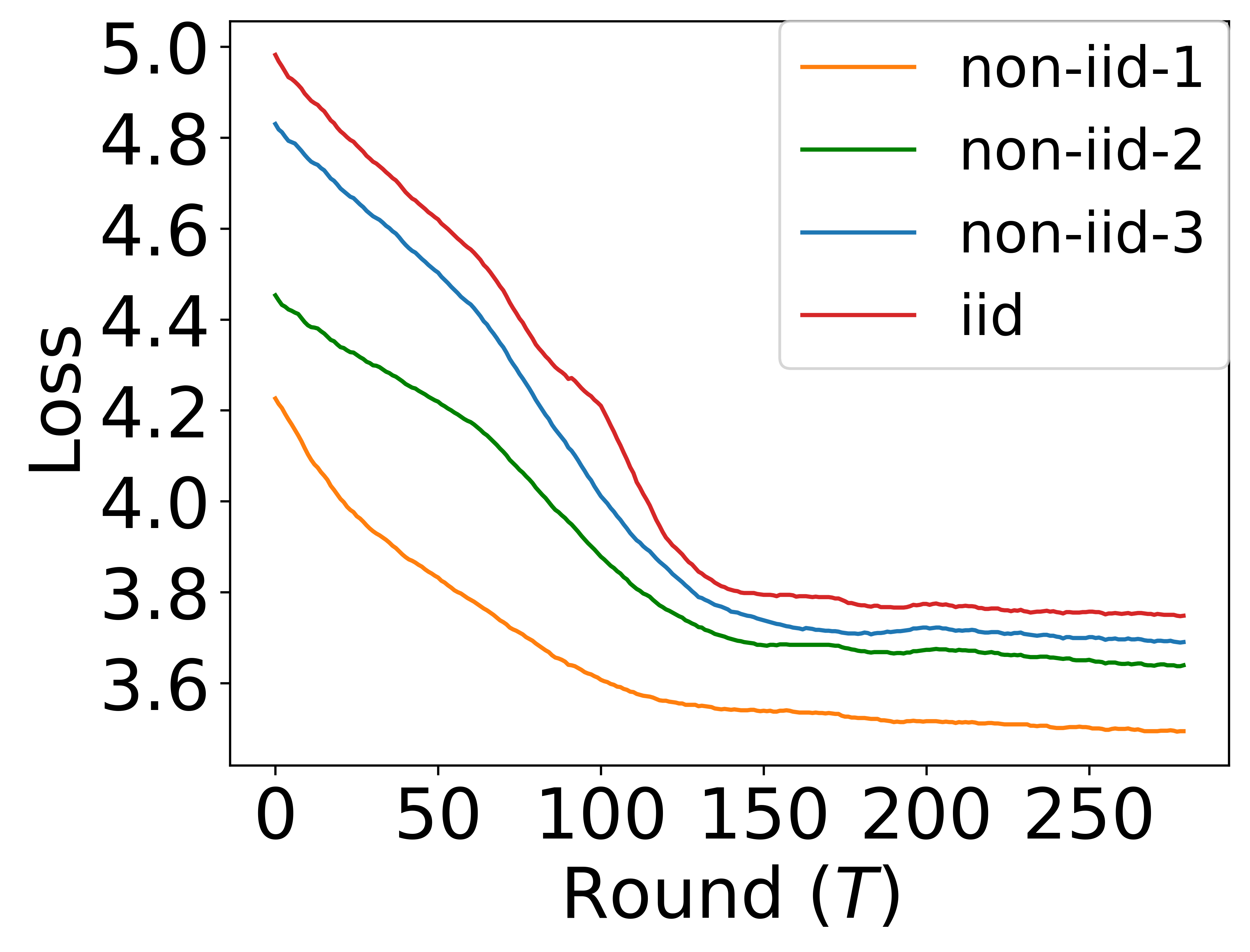}  
} 
\caption{$(a)$, $(b)$ and $(c)$ are our anomaly detection model's performance on four different data distribution(iid, non-iid-1, non-iid-2, non-iid-3) aganist Byzantine attacks(Gaussian noisy, sign flipping, same value).}     
\label{ours Vs abnormal all non-iid}     
\end{figure}
\par Figure \ref{ours Vs abnormal all non-iid} $(a)$, $(b)$ and $(c)$ show the influence of different data distributions on our detection model. For different data distributions, the detection ability of the model is different, but it is worth pointing out that : as the degree of non-iid of the data increases, the detection ability of the model increases instead. 

\subsubsection{\textbf{Impact of Unified Update}}
\begin{figure}[!h] \centering    
\subfigure[Same value] {
 \label{fig:a}     
\includegraphics[width=1.8in]{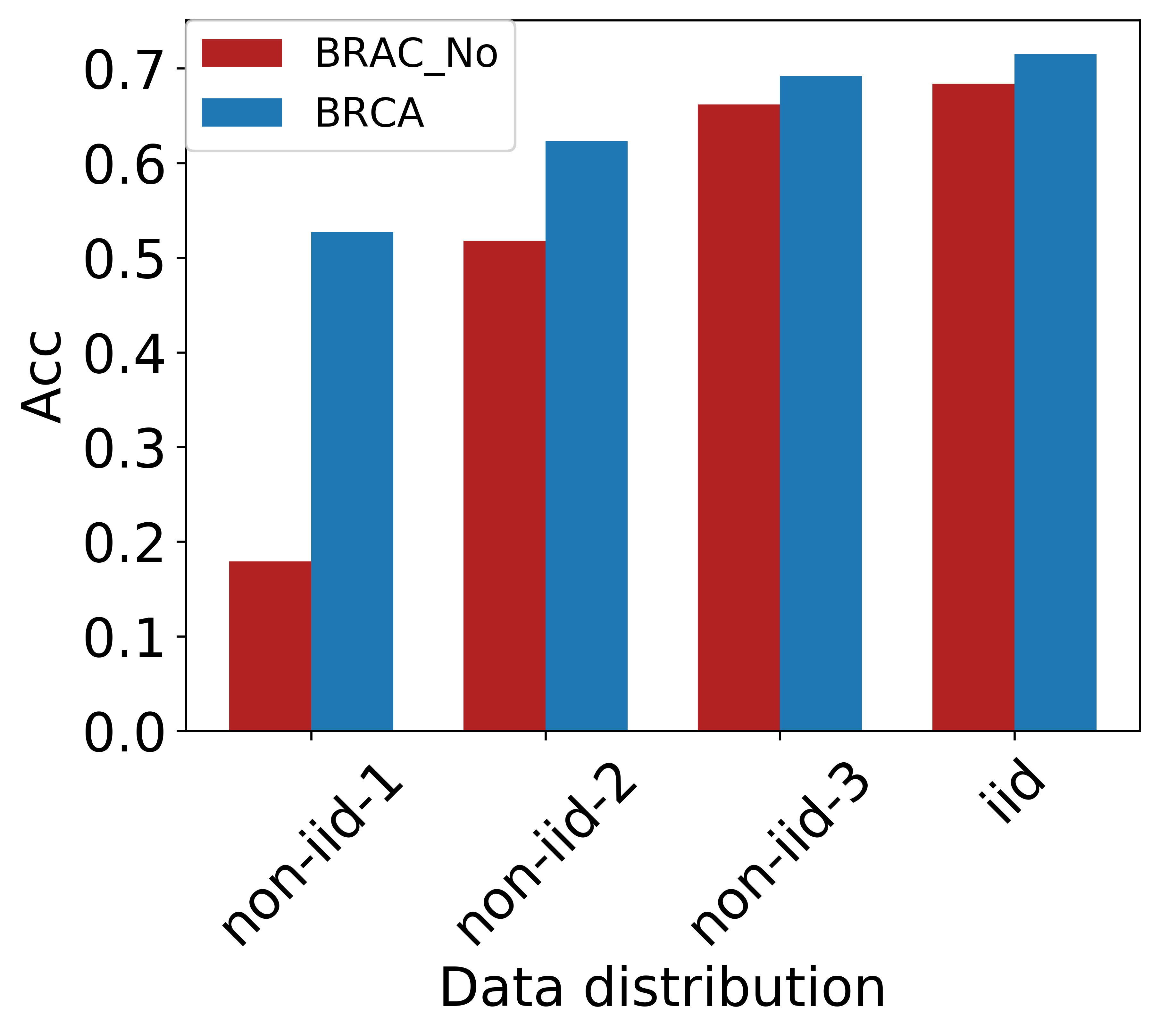}  
}     
\subfigure[Sign flipping] { 
\label{fig:b}     
\includegraphics[width=1.8in]{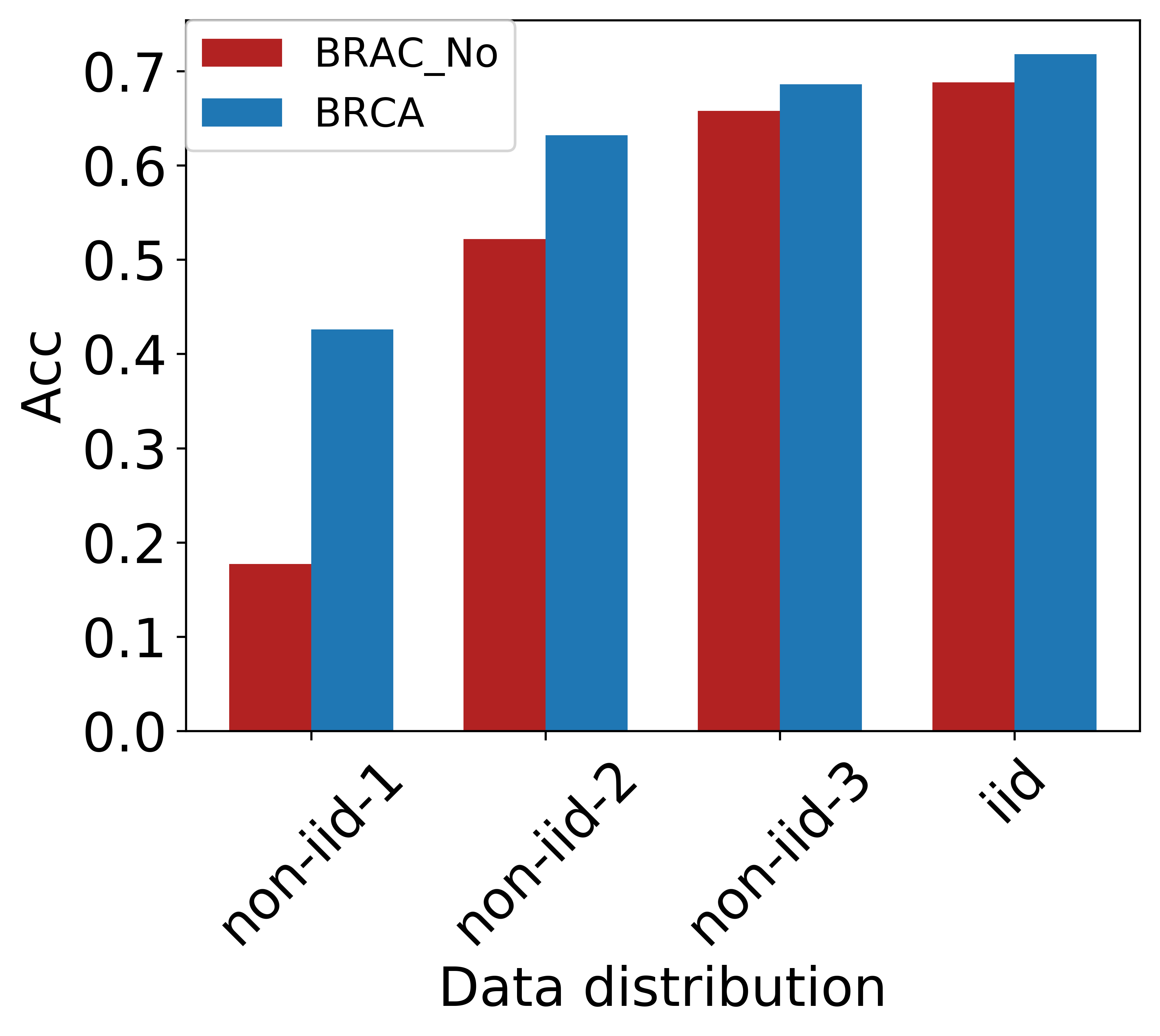}     
}    
\subfigure[Gaussian noisy] { 
\label{fig:b}     
\includegraphics[width=1.8in]{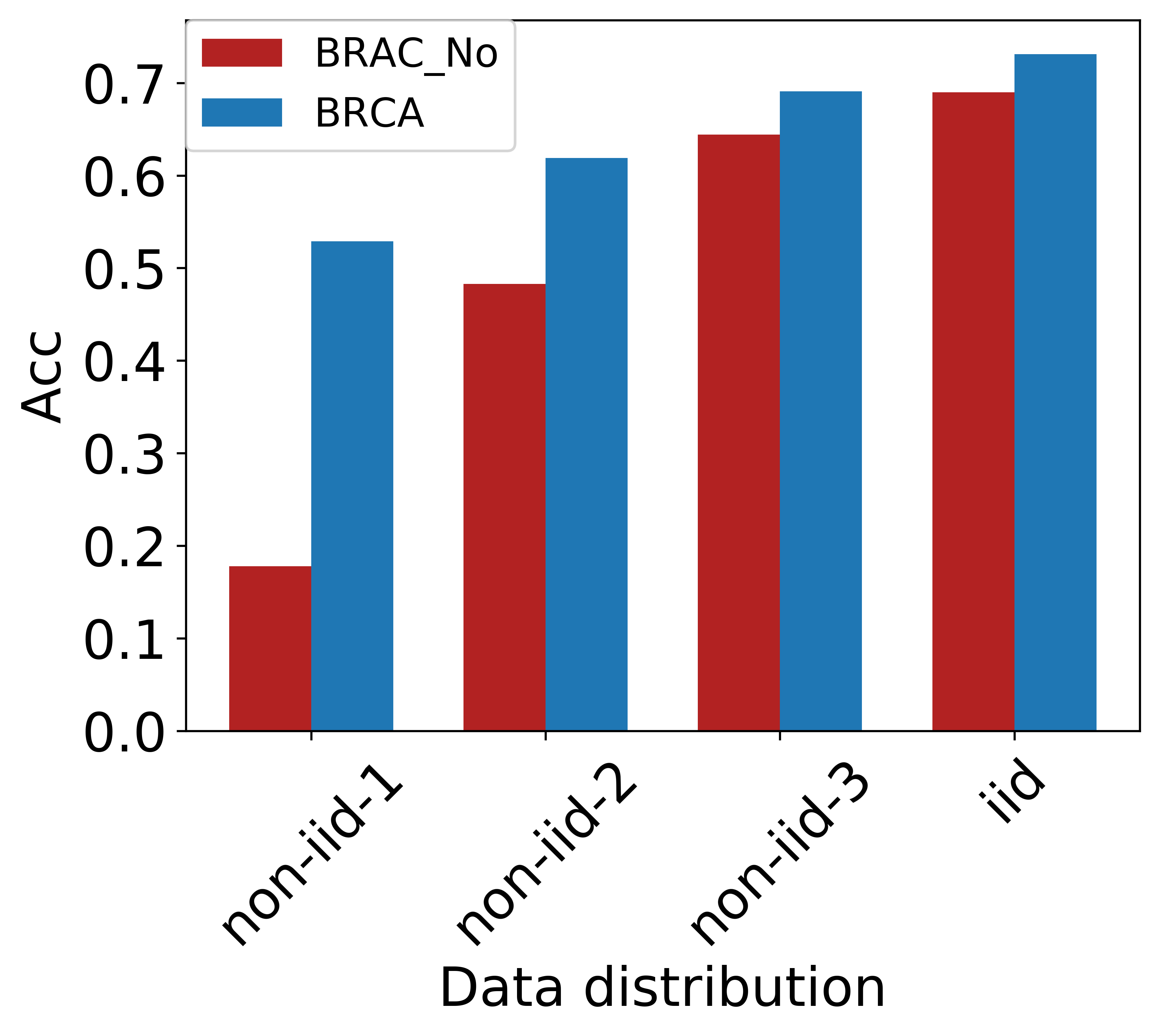}     
}   
\caption{The accuracy of BRCA and BRCA\_No on Cifar10. BRAC\_No is based on BRCA with unified update removed.}     
\label{BRCA and BRCA_No}     
\end{figure}

In this part, we study the impact of the unified update on the global model. Figure \ref{BRCA and BRCA_No} shows the accuracy of the global model with and without unified update on Cifar10. 

From non-iid-1 to iid, the improvement of the global model’s accuracy by unified update is as follows: $35.1\% \rightarrow 13.6\% \rightarrow 4.7\% \rightarrow 2.3\% $ (Same value), $34.8\% \rightarrow 10.5\% \rightarrow 3.0\% \rightarrow 3.1\%$ (Gaussian noisy), $24.9\% \rightarrow 9.9\% \rightarrow 2.8\% \rightarrow 3.0\%$ (Sign flipping). Combined with figure \ref{BRCA and BRCA_No}, it can be clearly found that the more simple the client data is, the more obvious the unified update will be to the improvement of the global model.  

When the data is non-iid, the directions of the model updates between clients are different. The higher the degree of non-iid of data, the more significant the difference. The global model obtained by weighted aggregation does not fit well with the global model. Unified update on the shared data can effectively unify the model updates of multiple clients, giving the global model a consistent direction.  

Therefore, it is necessary to implement a unified update to the primary aggregation model when data is non-iid.

\subsubsection{\textbf{Impact of Non-iid}}

\begin{table}[!h]
    \renewcommand\arraystretch{1.5}
    \scriptsize
	\centering
	\begin{tabular}{cccccccc}
	    \toprule
	     Attacks & \diagbox{Distri}{Defenses}  & No & Krum & GeoMed  & Abnormal & TrimmedMean &BRCA \\     
		\hline
		
		 \multirow{4}*{Same value} & Non-iid-1 & 0.1 & 0.1 & 0.1 & 0.178 & 0.1 & $\bm{0.529} $\\
		 
		 & Non-iid-2 & 0.101 & 0.207 & 0.205 & 0.483 & 0.1 & $\bm{0.619}$ \\
		 
		 & Non-iid-3 & 0.1 & 0.398 & 0.398 & 0.644 & 0.1  &  $\bm{0.691}$ \\

		 & iid & 0.098 & 0.696 & 0.705 & 0.690 & 0.101  & $\bm{0.713}$ \\
		\hline
		\multirow{4}*{Gaussian noisy} & Non-iid-1 & 0.143 & 0.1 & 0.1 & 0.179 & 0.110  & $\bm{0.527} $\\

		& Non-iid-2 & 0.1 & 0.191 & 0.204 & 0.518 & 0.059  & $\bm{0.623}$ \\

		& Non-iid-3 & 0.1 & 0.409 & 0.394 & 0.662 & 0.171 & $\bm{0.692}$ \\

		& iid & 0.1 & 0.697 & 0.694 & 0.684 & 0.120 & $\bm{0.715}$ \\
		\hline

		\multirow{4}*{Sign flipping} & Non-iid-1 & 0.1 & 0.101 & 0.1 & 0.177 & 0.1  & $\bm{0.426} $\\

		& Non-iid-2 & 0.1 & 0.192 & 0.214 & 0.522 & 0.1  & $\bm{0.621}$ \\

		& Non-iid-3 & 0.1 & 0.397 & 0.398 & 0.658 & 0.1  & $\bm{0.686}$ \\

		& iid & 0.1 & 0.697 & 0.703 & 0.688 & 0.1 & $\bm{0.718}$ \\
		\hline
	\end{tabular}
	\caption{The accuracy of the six defenses under four different data distributions on Cifar10, against three attacks.}
	\label{accuracy ,Cifar10}
\end{table}

\begin{table}[!h]
    \renewcommand\arraystretch{1.5}
    \scriptsize
	\centering
	\begin{tabular}{cccccccc}
	    \toprule
	     Attacks & \diagbox{Distri}{Defenses}  & No & Krum & GeoMed  & Abnormal & TrimmedMean &BRCA \\     
		\hline
		
		 \multirow{4}*{Same value} & Non-iid-1 & $2.84e^{16}$ & 11.72 & 9.61 & 2.29 & $6.05e^{17}$  & $\bm{2.09} $\\

		& Non-iid-2 & $6.99e^{16}$ & 7.29 & 8.01 & 2.06 & $3.63e^{16}$ &  $\bm{1.95}$ \\
		
		& Non-iid-3 & $4.48e^{16}$ & 2.35 & 2.38 & 1.893 & $3.37e^{16}$  & $\bm{1.83}$ \\
	
		& iid & $1.51e^{16}$ & 0.794 & 0.774 & 1.837 & $3.17e^{16}$  & $\bm{1.79}$ \\
		\hline
		
		\multirow{4}*{Gaussian noisy} & Non-iid-1 & $8.635e^{4}$ & 8.41 & 9.37 & 2.29 & 936.17  & $\bm{1.54} $\\
		
		& Non-iid-2 & 9.51 & 7.57 & 8.37 & 1.34 & 7.98 & $\bm{1.12}$ \\

		& Non-iid-3 & 8.22 & 2.01 & 2.31 & 0.94 & 6.07  & $\bm{0.857}$ \\

		& iid & 8.09 & 0.81 & 0.79 & 0.82 & 3.12 & $\bm{0.76}$ \\
		\hline

		\multirow{4}*{Sign flipping} & 	Non-iid-1 & 2.30 & 10.72 & 9.91 & 2.29 & 2.30  & $\bm{1.54} $\\
	
		& Non-iid-2 & 2.31 & 7.77 & 7.10 & 1.34 & 2.30  & $\bm{1.12}$ \\
	
		& Non-iid-3 & 2.31 & 2.36 & 2.13 & 0.94 & 2.30  & $\bm{0.848}$ \\
	
		& iid & 2.31 & 0.79 & 0.80 & 0.81 & 2.31  & $\bm{0.76}$ \\
		\hline
	\end{tabular}
	\caption{The loss of the six defenses under four different data distributions on Cifar10, against three attacks.}
	\label{loss , Cifar10}
\end{table}

Table \ref{accuracy ,Cifar10} and table \ref{loss , Cifar10}  show the accuracy and loss of each defense method under different data distributions on Cifar10. It can be seen that our method is the best, and the performance is relatively stable for different data distributions. The higher the degree of non-iid of data, the more single the data of each client, the lower the performance of the defense method.
\par Our analysis is as follows: 
1) The non-iid of data among clients causes large differences between clients' models. For the anomaly model, it is difficult for the defense method to judge whether the anomaly is caused by the non-iid of the data or by the Byzantine attacks, which increases the difficulty of defending the Byzantine attack to a certain extent. 
2) \emph{Krum} and \emph{GeoMed} use statistical knowledge to select the median or individual client's model to represent the global model. This type of method can effectively defend against Byzantine attacks when the data is iid. However when the data is non-iid, each client's model only focuses on a smaller area, and its independence is high, can not cover the domain of other clients, and naturally cannot represent the global model. 
3) \emph{Trimmed Mean} is based on the idea of averaging to defend against Byzantine attacks. When the parameter dimension of the model is low, it has a good performance. But as the complexity of the model increases, the method can not even complete convergence.

\section{Conclusion}
In this work, we propose a robust federated learning against Byzantine attacks when the data is non-iid.
\emph{BRCA} detects Byzantine attacks by \emph{credibility assessment}. At the same time, it makes the unified updating of the global model on the shared data, so that the global model has a consistent direction and its performance is improved.
\emph{BRCA} can make the global model converge very well when facing different data distributions. And for the pre-training of anomaly detection models, the transfer learning can help the anomaly detection model get rid of its dependence on the test data set.
Experiments have proved that \emph{BRCA} performs well both on non-iid and iid data, especially on non-iid, it is more prominent. 
In the future, we will improve our methods by studying how to protect the privacy and security of shared data.

\bibliography{acml19}

\end{document}